\documentclass[11pt]{article}
\usepackage[margin=1in]{geometry}
\usepackage{parskip}
\usepackage{authblk}
\usepackage{hyperref}

\usepackage[style=apa,natbib=true]{biblatex}
\usepackage{amsmath}
\usepackage{amssymb}
\usepackage{amsfonts}
\usepackage{amsthm}
\usepackage{mathtools}
\usepackage{dsfont}
\usepackage{stmaryrd}
\usepackage{nicefrac}
\usepackage{microtype}
\usepackage{xcolor}
\definecolor{FLgreen}{HTML}{1b7a5a}
\definecolor{FLrust}{HTML}{b8430f}
\usepackage{tcolorbox}
\usepackage{algpseudocode}
\usepackage{algorithm}
\usepackage{tikz}
\usepackage{tikz-cd}
\usepackage{bussproofs}
\usepackage{listings,lstautogobble}
\usepackage{caption}
\usepackage{subcaption}
\usepackage{float}
\usepackage{appendix}
\usepackage{multicol}
\usepackage{multirow}
\usepackage{tabularx}
\usepackage{booktabs}
\usepackage[percent]{overpic}
\usepackage[normalem]{ulem}
\usepackage{cleveref}
\usepackage{comment}
\usetikzlibrary{calc,positioning,arrows.meta,shapes}

\newcommand{\td}[1]{}

\hypersetup{colorlinks,citecolor=FLgreen,linkcolor=FLrust,urlcolor=FLrust}

\renewcommand{\epsilon}{\varepsilon}
\renewcommand{\phi}{\varphi}

\renewcommand{\geq}{\geqslant}

\title{Exploring Flow-Lenia Universes with a Curiosity-driven AI Scientist:\\ Discovering Diverse Ecosystem Dynamics}

\author[1,*]{Thomas Michel}
\author[2,*]{Marko Cvjetko}
\author[2]{Gautier Hamon}
\author[2]{Pierre-Yves Oudeyer}
\author[2,3]{Cl\'{e}ment Moulin-Frier}
\affil[1]{Univ. Lille, Inria, CNRS, Centrale Lille, CRIStAL, France}
\affil[2]{Inria Center at the University of Bordeaux, France}
\affil[3]{Inria, INSA Lyon, CITI, UR3720, 69621 Villeurbanne, France}
\affil[*]{These authors contributed equally. Correspondence: \texttt{marko.cvjetko@inria.fr}.}
\date{}

\begin{document}
\maketitle

\begin{abstract}
    We present a curiosity-driven AI scientist method for discovering system-level dynamics in Flow-Lenia, a continuous cellular automaton (CA) with mass conservation and parameter localization. Building on prior work that uses diversity search in Lenia to find individual self-organized patterns, we adapt Intrinsically Motivated Goal Exploration Processes (IMGEPs) to large environments of interacting patterns, using simulation-wide metrics such as evolutionary activity, compression ratio, and multi-scale matter distribution. We apply IMGEP in two exploration experiments: one targeting ecosystem-level dynamics, the other matter movement through obstacle-laden environments. In both, IMGEP illuminates significantly more of the metric space than random search and reveals self-organized behaviors qualitatively resembling many biological phenomena. Leveraging the resulting archive, we then run a scaling study across six spatial scales and seven time horizons, uncovering macro-scale organization with no analogue at the base scale and characterizing how goal-space metrics behave at scale. This illustrates a strength of our approach: a relatively cheap large-scale diversity search can act as a principled scaffold for designing subsequent, more expensive experiments, enabling an iterative loop of experiment design, inspection, and redesign --- supported by an interactive exploration tool that keeps scientists in the loop. Though demonstrated with Flow-Lenia, this approach potentially applies to other parameterizable complex systems where studying bottom-up collective behavior is of interest.

\end{abstract}

\makeatletter
\def\@makefnmark{}
\footnotetext{This paper features a companion website with a visualization tool that allows to explore a subset of the Flow-Lenia environments explored in the experiments of this paper. The website is available at
    \footnote{Companion site: \url{https://developmentalsystems.org/Flow-Lenia-Universes-Journal/}.}

    \url{https://developmentalsystems.org/Flow-Lenia-Universes-Journal/}}
\def\@makefnmark{\hbox{\@textsuperscript{\normalfont\@thefnmark}}}
\makeatother

\begin{figure}[tp]
    \centering
    \includegraphics[width=\textwidth]{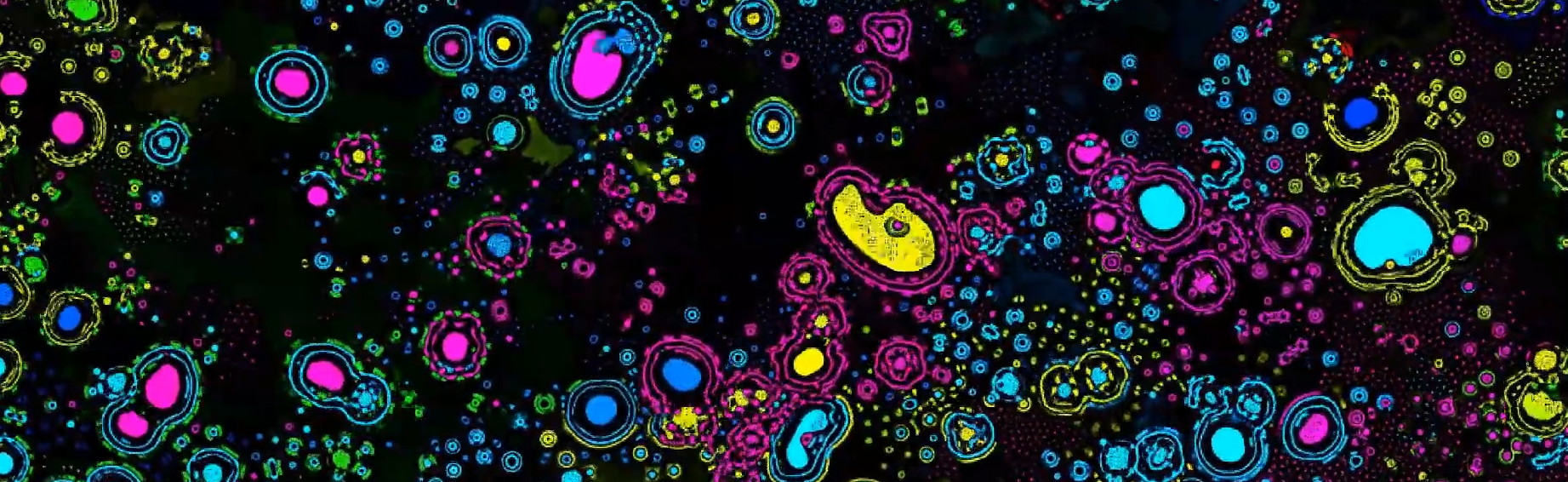}
    \caption{Snapshot of an advanced state of Flow-Lenia. Each color represents a different set of localized parameters, determining how matter behaves. Regions with brighter colors represent higher concentrations of mass. In this paper, we present an AI scientist method, based on IMGEPs, to systematically discover novel dynamics in large-scale Flow-Lenia simulations.}
    \label{fig:flow_lenia_example}
    \vspace{-5pt}
\end{figure}

\section{Introduction}

Understanding how complex, system-level properties emerge in self-organizing systems represents a fundamental challenge across many scientific fields. From ecological dynamics to open-ended evolution, these phenomena arise from interactions between multiple entities over time, making them difficult to discover through traditional exploration methods that focus on individual patterns or behaviors.

Searching for emergent ecosystem dynamics in artificial systems entails two main challenges, the first being the simulation of a dynamical system with the potential to self-organize complex interactions and emergent evolution under certain conditions. Computational models offer controlled environments to investigate such emergent properties, with cellular automata (CAs) serving as powerful substrates for exploring self-organization and complexity \citep{langton1995artificial, Beer2014Cognitive}. In this paper, we rely on Flow-Lenia \citep{plantec2023flow}, a recent continuous CA integrating mass conservation and parameter localization. This approach enables the simulation of multiple interacting patterns within a shared environment, creating the potential for emergent evolution and system-wide phenomena such as those shown in Figure \ref{fig:flow_lenia_example}.
\begin{figure}[tp]
    \centering
    \resizebox{\textwidth}{!}{
        \begin{tikzpicture}[
                node distance=1.2cm,
                box/.style={rectangle, draw, rounded corners,
                        minimum width=3.4cm, minimum height=0.9cm,
                        align=center, font=\footnotesize, fill=blue!10},
                arrow/.style={->, >=stealth, thick, draw=blue!60},
                section/.style={rectangle, text width=4.6cm, minimum height=1cm,
                        align=center, font=\small\bfseries,
                        fill=blue!20, rounded corners, inner sep=5pt},
                mlabel/.style={align=center, font=\footnotesize\bfseries,
                        text width=4.4cm},
                mbox/.style={rectangle, rounded corners, fill=green!10,
                        inner sep=8pt, align=left, font=\footnotesize,
                        text width=4.2cm},
                ibox/.style={rectangle, rounded corners, fill=orange!10,
                        inner sep=8pt, align=center, font=\footnotesize,
                        text width=4.4cm},
                img/.style={inner sep=0, outer sep=0}
            ]
            \useasboundingbox (-2.7,-3) rectangle (14.5,5.8);

            \node[section] at (0,5)  {IMGEP for Ecosystem\\Exploration};
            \node[section] at (6,5)  {Multi-metric\\Evaluation};
            \node[section] at (12,5) {Discovered System-wide\\Dynamics};

            \node[box, fill=blue!10] (goal)    at (0, 3.6) {1. Sample New Goal in\\System-wide Metrics Space};
            \node[box, fill=blue!15] (select)  at (0, 2.3) {2. Select Parameters from\\Previous Similar Simulation};
            \node[box, fill=blue!20] (mutate)  at (0, 1.0) {3. Mutate Simulation\\Parameters};
            \node[box, fill=blue!25] (run)     at (0,-0.3) {4. Run Flow-Lenia\\Simulation};
            \node[box, fill=blue!30] (compute) at (0,-1.6) {5. Compute Simulation\\Metrics \& Archive};
            \draw[arrow] (goal)   -- (select);
            \draw[arrow] (select) -- (mutate);
            \draw[arrow] (mutate) -- (run);
            \draw[arrow] (run)    -- (compute);
            \draw[arrow] (compute.west) .. controls +(-0.8,0) and +(-0.8,0) .. (goal.west);

            \node[mlabel] (top_lbl) at (6, 3.7) {System-level Metrics:};
            \node[mbox, below=0.15cm of top_lbl] (top_box) {%
                - Evolutionary Activity\\(meaningful adaptation)\\
                - Compression Complexity\\(pattern diversity)\\
                - Multi-scale Entropy\\(spatial organization)};
            \node[mlabel, below=0.5cm of top_box] (bot_lbl) {Parameters Explored:};
            \node[mbox, below=0.15cm of bot_lbl] {%
                - Kernel weights\\
                - Growth functions\\
                - Initial distribution of matter};

            \node[img] at (10.85, 3.0) {\includegraphics[width=2cm,height=2cm]{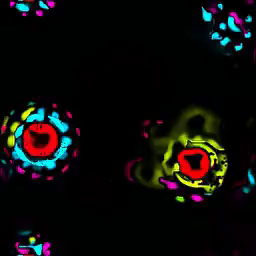}};
            \node[img] at (13.15, 3.0) {\includegraphics[width=2cm,height=2cm]{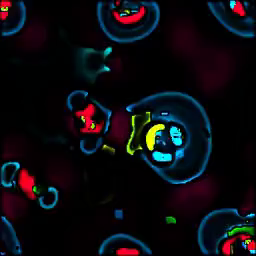}};
            \node[img] at (10.85, 0.7) {\includegraphics[width=2cm,height=2cm]{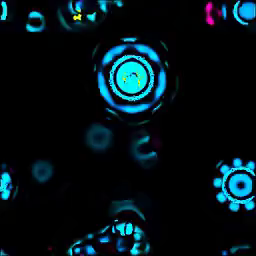}};
            \node[img] at (13.15, 0.7) {\includegraphics[width=2cm,height=2cm]{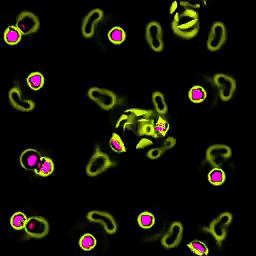}};
            \node[ibox] at (12,-1.9) {Discovery of diverse system-level dynamics not found through random exploration};
        \end{tikzpicture}}
    \caption{IMGEP approach for exploring a diversity of self-organization in Flow-Lenia. The algorithm samples new goals in a system-wide metric space, selects parameters from previous similar simulations, mutates them, runs simulations, and computes metrics to guide further exploration. This process discovers a variety of dynamics, including dense matter clusters akin to colonies, behavior resembling allopatric speciation, feeding, and more.}
    \label{fig:imgep_overview}
    \vspace{-5pt}
\end{figure}
The vast parameter spaces of these complex systems bring a second challenge: how can researchers efficiently discover diverse, interesting dynamics? Random or exhaustive sampling proves ineffective in large parameter spaces, often undersampling low-volume regions that might support interesting behaviors, while manual human exploration cannot systematically cover high-dimensional spaces \citep{etcheverry2020hierarchically,etcheverry2023curiosity}. Diversity search algorithms have been proposed as efficient exploration methods in such systems and have recently been applied to explore large rule spaces of parameterizable CAs \citep{hamon2024discovering,faldor2024toward}. In this paper, we rely on Intrinsically Motivated Goal Exploration Processes (IMGEPs) \citep{forestier2022intrinsically}, a family of autotelic diversity search algorithms: IMGEP algorithms self-generate and try to achieve diverse goals to explore a space of behaviors, cf. left side of Figure \ref{fig:imgep_overview}. IMGEPs have been shown to drive efficient exploration of vast spaces of behaviors in multiple domains spanning robotics \citep{cully2015robots, forestier2022intrinsically}, biological networks \citep{etcheverry2024ai}, code \citep{wang2023voyager, faldor2024omni,pourcel2024aces}, within the broader family of so-called "AI scientists" systems \citep{lu2024ai,gottweis2025towards}. As they include a general definition of goals as abstract reward functions subject to arbitrary constraints, IMGEPs include as a special instance quality-diversity algorithms used in the artificial life community \citep{faldor2024toward}.
Previous works have shown how IMGEPs enable the discovery of diverse patterns in continuous cellular automata \citep{reinke2020a, etcheverry2020hierarchically}, including emergent sensorimotor agents \citep{hamon2024discovering}. However, these approaches focused on discovering individual patterns (spatially localized or Turing patterns). Here we apply it for the first time in Flow-Lenia in order to explore a space of simulation-wide metrics that capture aspects of evolutionary dynamics and complexity across multiple scales.

We demonstrate how this approach can systematically discover diverse system-level dynamics that would remain hidden under random search. We modified aspects of Flow-Lenia to better support multi-species evolution, creating a testbed in which multiple patterns governed by different update rules can coexist and interact. We apply IMGEP in two complementary settings. In the first, we target ecosystem-level behavior through a goal space combining evolutionary activity, compression ratio, and multi-scale entropy; the resulting discoveries suggest ecological interactions such as feeding, tightly packed clumps of matter resembling colonies, and self-organized patterns of different scales coexisting in a shared environment (Figures 3 and 8a). In the second, we constrain the environment with obstacles and use the search process to find different spatial distributions of matter at the end of the simulation; here IMGEP uncovers varied movement regimes including directed transport, fragmentation into faster-moving entities that navigate narrow corridors, and dynamics reminiscent of allopatric speciation.
Beyond direct discovery, the archive built by IMGEP serves as a principled scaffold for designing subsequent, larger-scale experiments. We illustrate this by selecting a diverse subset of discovered universes (using farthest point sampling in the z-scored goal space) and rerunning them across six spatial scales and seven time horizons. This scaling study uncovers macro-scale organization with no analogue at the base scale — coherent structures larger than entire base-sized grids — and characterizes how each goal-space metric behaves under spatial and temporal scaling. We highlight the potential to use our method iteratively: a relatively cheap small-scale diversity search illuminates a broad swath of parameter space, inspection of the resulting archive identifies which directions are worth investigating at scale, and the scaled experiments in turn reshape the design of subsequent searches.
We support this loop with an interactive exploration tool that bridges automated discovery and human analysis, letting researchers traverse the archive, inspect simulations at intermediate steps, and identify promising regions of parameter space. The resulting human-AI collaborative workflow allows scientists to investigate the system without being limited by low-level manual parameter tuning. Although we demonstrate this approach on Flow-Lenia, it potentially applies to any parameterizable complex system for which relevant metrics can be computed.

\begin{figure}[h]
    \centering
    \begin{tabular}{cc}
        \includegraphics[width=0.45\columnwidth]{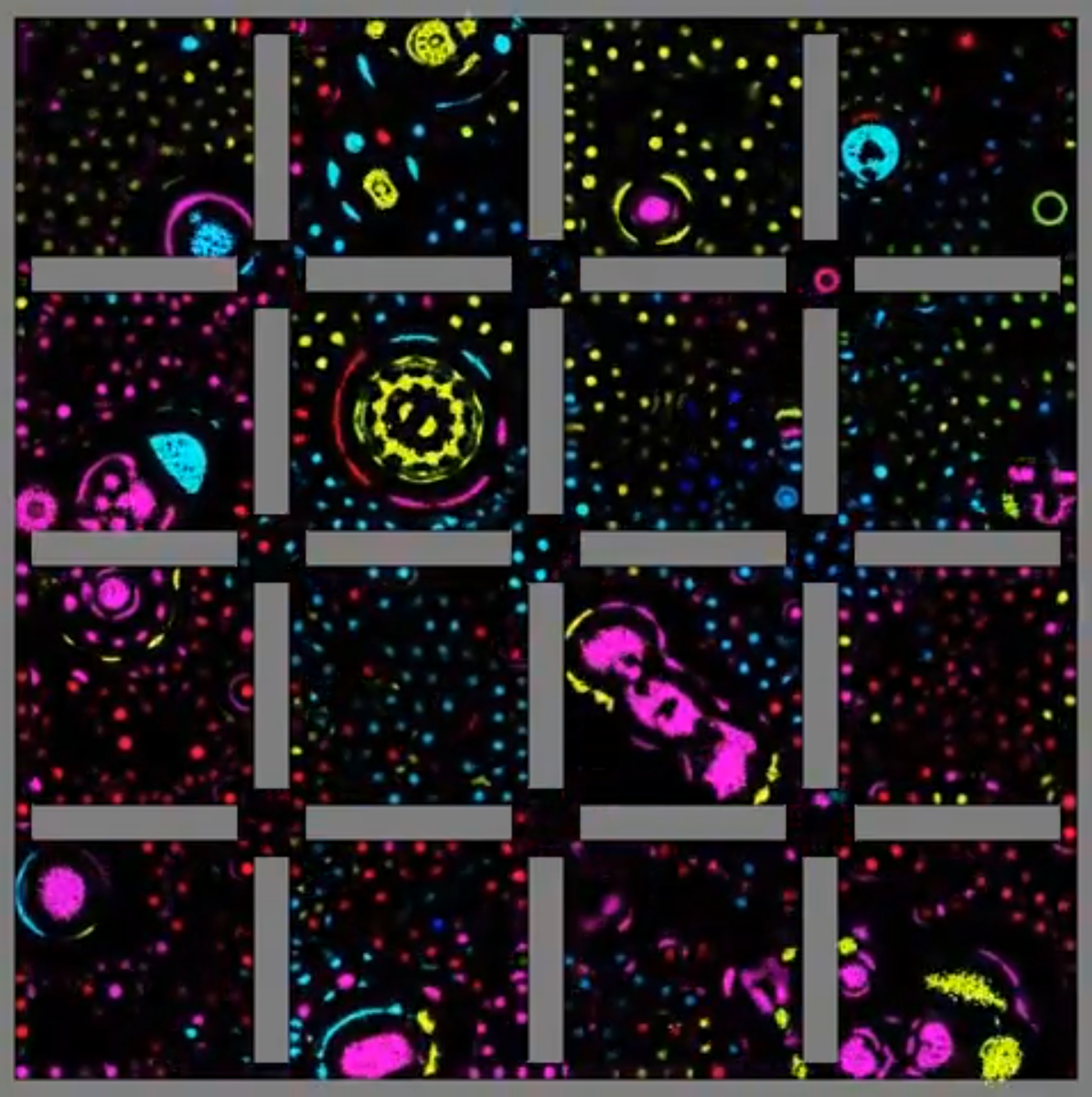} &
        \includegraphics[width=0.45\columnwidth]{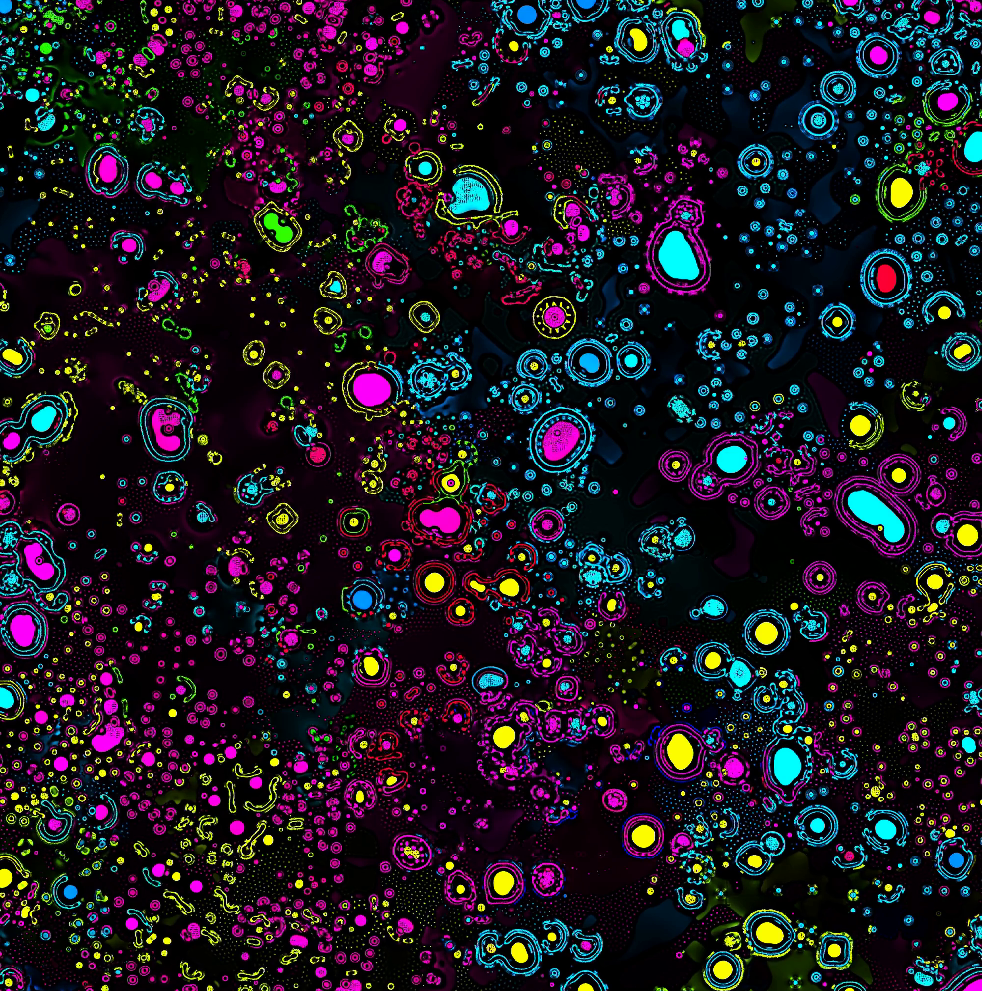}   \\
    \end{tabular}
    \caption{Examples of environments with multiple parameters co-existing and interacting in Flow-Lenia. Gray bars represent walls blocking the flow of matter, creating environmental constraints that influence the system's dynamics.}
    \label{fig:ecological}
\end{figure}

\section{Related Works}

Our research intersects open-ended evolution, cellular automata, and intrinsically motivated exploration for automated scientific discovery, building upon existing work in these areas.

\subsection{Open-Ended Evolution in Artificial Systems}

Creating artificial systems capable of open-ended evolution remains a fundamental challenge in ALife. \citet{bedau1996} established quantitative measures of evolutionary activity that have become foundational for evaluating evolutionary dynamics. \citet{taylor2015requirements} and \citet{soros2014identifying} identified requirements for open-ended evolution, including large state spaces, ecological interactions, and hierarchical organization. Despite these advances, most artificial evolutionary systems reach complexity plateaus \citep{stanley2019open}.

The evaluation of open-endedness has evolved from early parameters like Langton's $\lambda$ \citep{langton1990computation} and Wolfram's classifications \citep{wolfram1984universality} to more comprehensive metrics, such as those in the MODES toolbox \citep{dolson2019modes} and compression-based approaches \citep{zenil2015two}, which measure diversity, complexity, and evolvability in different ways.

\subsection{Cellular Automata and Flow-Lenia}

Cellular automata have long served as testbeds for studying open-ended evolution, from Conway's Game of Life \citep{adamatzky2010game} to Langton's self-replicating loops \citep{langton1984self} and Beer's exploration of autopoiesis \citep{beer2004}. Sayama's work on evolving ecosystems \citep{sayama1999} and genetic evolution in CAs \citep{salzberg2004} pioneered approaches for encoding and propagating information.

Lenia \citep{chan2018lenia} generalized the Game of Life to the continuous domain, generating diverse life-like patterns. Several variants aim to promote open-ended evolution, such as Chan's model with localized parameters (akin to genes) \citep{chan2023large}, and Flow Lenia~\citep{plantec2023flow}, which, in addition to local parameters (i.e., local update rules), also incorporates mass conservation, shown to increase evolutionary activity \citep{hickinbotham2015conservation}.

\subsection{Intrinsically Motivated Exploration for Automated Scientific Discovery}

Random sampling is inadequate to efficiently map parameter space to system's behaviors in complex systems, due to the high dimensionality of the parameter space and the strong non-linearity of the parameter-to-behavior mapping \citep{etcheverry2020hierarchically}. Approaches based on computational models of curiosity or intrinsically motivated exploration \citep{oudeyer2007intrinsic}, related to novelty search methods \citep{lehman2011a}, can be used for automated scientific discovery \citep{etcheverry2020hierarchically, etcheverry2024ai}, within the large family of AI scientist approaches \citep{lu2024ai,gottweis2025towards}.
The IMGEP framework \citep{forestier2022intrinsically} implements a form of intrinsically motivated exploration that efficiently discovers diverse behaviors: it enables to build agents that can self-generate, self-select and pursue their own goals, i.e. autotelic agents~\citep{colas2022autotelic}. They have demonstrated to be effective approaches for discovering diverse patterns in self-organizing artificial and biological systems \citep{reinke2020a,etcheverry2020hierarchically,etcheverry2024ai}, in physical systems \citep{falk2024curiosity}, or in chemistry \citep{grizou2020curious}. These methods, as well as related methods like novelty search \citep{kumar2024automating} or quality-diversity \citep{faldor2024toward}, have proven valuable for exploring self-organization of sensorimotor 'agents' in CA systems such as Lenia \citep{hamon2024discovering}. However, they have not previously been applied to investigate evolutionary and/or ecosystemic dynamics in large CA environments. 

\subsection{Measuring Complexity and Evolutionary Activity}

Defining and measuring complexity has proven to be a difficult task, with many different metrics proposed over the years \citep{Mitchell2010Complexity}. Complexity measures based on information theory \citep{zenil2015two, cisneros2019evolving} connect to Kolmogorov complexity, while concepts like thermodynamic depth \citep{lloyd1988complexity} quantify the difference between a system's fine-grained and coarse-grained entropy. \citet{droop2012} introduced non-neutral evolutionary activity metrics suited for systems with intrinsic fitness and distinct genotypes, measuring meaningful adaptive changes while filtering out neutral drift. 

We combine complexity and evolutionary activity metrics to discover Flow-Lenia universes that exhibit complex patterns and support long-term evolutionary trajectories through local parameter mutations. Our approach is based on IMGEP, which helps us systematically illuminate the multi-dimensional metric space, enabling the discovery of new universes and the study of conditions that foster open-ended evolution.

\section{Methods}

This section outlines our approach to investigating system-level dynamics in Flow-Lenia. We detail the automaton’s core components, including its parameter propagation mechanisms, and introduce system-wide metrics designed to characterize ecosystem-like environments. The section concludes with a description of our IMGEP implementation.

\subsection{Flow-Lenia}

Flow-Lenia extends the continuous cellular automaton Lenia by incorporating principles of mass conservation \citep{plantec2023flow}. While Lenia's state space is confined to the unit range $[0, 1]^C$, Flow-Lenia expands this to $\mathbb{R}_{\geq 0}^C$, where $C$ is the number of channels. This extension allows for unbounded positive real values, reflecting the system's interpretation of cell states as matter density rather than abstract activation levels.

The core components of Flow-Lenia include:

\begin{itemize}
    \item \textbf{State space}: $A^t: \mathcal{L} \to \mathbb{R}_{\geq 0}^C$ represents the distribution of matter across the grid $\mathcal{L}$ over $C$ different channels
    \item \textbf{Convolution kernels}: $K = \{K_i: \mathcal{L} \to [0, 1] \mid i = 1, \ldots, |K|\}$ define the range and strength of interactions between cell states.
    \item \textbf{Growth functions}: $G = \{G_i: [0,1] \to [-1,1] \mid i = 1, \ldots, |K|\}$ determine how interactions influence matter movement.
\end{itemize}

The system's dynamics unfold in two steps. First, an affinity map $U^t$ is computed for each channel $j$:

\begin{equation}
    U^t_j(x) = \sum_{i=1}^{|K|} h_i \cdot G_i(K_i * A^t_{c^i_0})(x) \cdot [c^i_1 = j]
\end{equation}

where $h_i \in \mathbb{R}$ weights the contribution of each kernel-growth function pair, $K_i * A^t_{c^i_0}$ is the convolution of kernel $K_i$ with its source channel $c^i_0$, and $[c^i_1 = j]$ is the Iverson bracket, equaling 1 when $c^i_1 = j$ and 0 otherwise. $c_0^i$ (resp. $c_1^i$) denotes the source (resp. target) channel of kernel $K_i$.

Second, a flow field $F^t$ is derived from this affinity map, which determines how matter will move:

\begin{equation}
    \begin{cases}
        F_i^t = (1-\alpha^t)\nabla U_i^t - \alpha^t \nabla A_\Sigma^t \\
        \alpha^t(p) = [(A_\Sigma^t(p)/\theta_A)^n]_0^1
    \end{cases}
\end{equation}

This flow combines an attraction toward high-affinity regions with a diffusion effect (modulated by $\theta_A$) that prevents excessive matter concentration.

To move matter according to the computed flow field while preserving mass conservation, Flow-Lenia uses a reintegration tracking method \citep{moroz2020reintegration}. This approach works like a grid-based particle system where each cell sends matter according to the flow field. When matter from a source cell flows to a destination that overlaps multiple grid cells, the matter is distributed proportionally across these cells. This ensures that the total amount of matter in the system remains constant, a crucial property that distinguishes Flow-Lenia from traditional cellular automata.

\subsection{Parameter Localization and Mixing Rules}

Since Flow-Lenia's computation literally displaces individual cells of matter, it enables one to attach any information to pieces of matter. For instance, it is possible to attach parameters of update rules, i.e., functions which determine the next state of a cell. These parameters will flow with matter, enabling the co-existence of distinct patterns, each driven by different rules, within a shared grid. We define a parameter map $P: \mathcal{L} \to \Theta$, where $\Theta$ is the parameter space. In our implementation, we embed the kernel weight vector $h \in \mathbb{R}^{|K|}$, modifying the affinity map computation:

\begin{equation}
    U^t_j(x) = \sum_{i=1}^{|K|} P^t_i(x) \cdot G_i(K_i * A^t_{c^i_0})(x) \cdot [c^i_1 = j]
\end{equation}

\subsubsection{The Role of Mixing Rules}
\label{sec:mixing_rules}

As matter flows across the grid, streams carrying different parameters can converge on the same destination cell. The system must resolve which parameters the destination cell will adopt. We refer to the methods for resolving these conflicts as \emph{mixing rules}. Formally, a mixing rule $\mathcal{M}$ determines the new parameters for a cell based on the incoming matter and parameters:

\begin{equation}
    P^{t+dt}(x_\text{dest}) = \mathcal{M}({(A^t(x_\text{src}), P^t(x_\text{src}), I(x_\text{src}, x_\text{dest})) \mid x_\text{src} \in \mathcal{L}})
\end{equation}

where $I(x_\text{src}, x_\text{dest})$ denotes the proportion of matter flowing from source cell $x_\text{src}$ to destination cell $x_\text{dest}$.

The choice of mixing rule shapes the evolutionary dynamics of the system. Different rules affect how parameters propagate when matter streams merge: some preserve distinct parameter sets, others blend them, and still others select based on local conditions. These cell-level decisions compound over time, determining whether the system supports sustained evolutionary activity or converges to homogeneous states.

\subsubsection{Alternative Mixing Rules}

We define below several existing mixing rules proposed in prior work, before introducing the negotiation rule.

\textbf{Weighted Average.} Computes new parameters as an average of incoming parameters, weighted by their associated matter quantities:
\begin{equation}
    P^{t+dt}(x_\text{dest}) = \frac{\sum_{x_\text{src}} A^t(x_\text{src}) I(x_\text{src}, x_\text{dest}) P^t(x_\text{src})}{\sum_{x_\text{src}} A^t(x_\text{src}) I(x_\text{src}, x_\text{dest})}
\end{equation}
This rule creates smooth parameter transitions but tends to homogenize the parameter space over time, reducing diversity.

\textbf{Stochastic Sampling.} Randomly selects one set of incoming parameters, with probabilities proportional to their associated matter quantities. This is the original approach proposed by \citet{plantec2023flow}:
\begin{equation}
    \mathbb{P}\left[P^{t+dt}(x_\text{dest})=P^t(x_\text{src})\right] = \frac{A^t(x_\text{src}) I(x_\text{src}, x_\text{dest})}{\sum_{x} A^t(x) I(x, x_\text{dest})}
\end{equation}
Stochastic sampling maintains parameter diversity but can lead to abrupt changes in local behavior.

\textbf{Stochastic Gene-wise Sampling.} Independently samples each parameter dimension from the incoming sets, potentially creating novel parameter combinations. This can generate new parameter vectors not present in any source cell, but may disrupt functional patterns that depend on specific parameter correlations.

\textbf{Softmax Variants.} Both the weighted average and stochastic sampling rules can be modified with a softmax temperature parameter $\tau$ that controls the influence of matter quantities on the selection. Higher temperatures yield more uniform selection; lower temperatures concentrate selection on the dominant contributor.

\textbf{Dot Product-based Selection.} Selects parameters based on their similarity to other incoming parameters, measured by dot products. This rule can promote either convergence to similar parameters (when selecting similar) or maintenance of distinct parameter regions (when selecting dissimilar).

\subsubsection{Negotiation Rule}

We introduce a ``negotiation rule'' that considers both the quantity of incoming matter and its affinity for the destination environment. The probability of a destination cell adopting parameters from a particular source is:

\begin{equation}
    \mathbb{P}\left[P^{t+dt}(x_\text{dest})=P^t(x_\text{src})\right] = \frac{e^{\beta A^t(x_\text{src}) I(x_\text{src}, x_\text{dest}) V^t(x_\text{src})}}{\sum_{x \in \mathcal{L}} e^{\beta A^t(x) I(x, x_\text{dest}) V^t(x)}}
\end{equation}

where $\beta$ is an inverse temperature parameter controlling selection pressure, and $V^t(x_\text{src})$ is an affinity map computed for the mixing step:

\begin{equation}
    V^t(x_\text{src}) = \sum_{j=1}^C \sum_{i=1}^{|K|} Q^t_i(x_\text{src}) \cdot G_i(K_i * A^t_{c^i_0})(x_\text{src}) \cdot [c^i_1 = j]
\end{equation}

Here, $Q^t_i(x)$ is a separate set of parameters used only for the mixing affinity computation.

The negotiation rule allows parameters to persist in regions where they achieve high affinity, even when competing against larger quantities of matter carrying different parameters. Parameters can therefore specialize for particular local environments, adapting to specific matter distributions or densities, without being overwhelmed by sheer quantity. We selected the negotiation rule based on experiments showing it produces the highest evolutionary activity among the rules we tested (Section \ref{sec:mixing_experiments}).

To promote diversity and evolution, we implement a mutation mechanism at a specified frequency, randomly selecting areas of the grid and applying multivariate Gaussian noise to the parameter map attached to matter.

\begin{equation}
    P^{t+dt}(x) = P^t(x) + \epsilon, \quad \epsilon \sim \mathcal{N}(0, \Sigma)
\end{equation}

\subsection{System-level Metrics}
\label{metrics}

To quantify emergent dynamics in Flow-Lenia, we use the following set of metrics:

\subsubsection{Non-neutral Evolutionary Activity}

The non-neutral evolutionary activity metric \citep{droop2012}, already used in a previous work on Flow-Lenia by \citet{plantec2023flow}, measures population changes in evolutionary systems while filtering out neutral drift:

\begin{equation}
    EA = \sum_{i}\sum_{t=1}^{T}\Delta_i(t)
\end{equation}

where $i$ indexes all components (species or genomes) that existed during the simulation, $T$ is the total number of time steps, and:

\begin{equation}
    \Delta_i(t) =
    \begin{cases}
        (p_i(t) - p_i(t-1))^2 & \text{if } p_i(t) > p_i(t-1) \\
        0                     & \text{otherwise}
    \end{cases}
\end{equation}

In our case, $p_i(t)$ represents the proportion of simulated mass carrying a local parameter $i$ at time $t$. We only consider identical local parameters to be the same "genome" type.  In practice, we sample $p_i(t)$ at a fixed interval $\Delta t$ rather than at every simulation step, and $\Delta_i$ is accumulated only across these sampled points, so reported EA values depend on the choice of $\Delta t$ (see Section~\ref{sec:mixing_experiments}).

\subsubsection{Compression-based Complexity Metric}

We use the size of MP4 video files encoding the simulation as a proxy for complexity, inspired by the concept of Kolmogorov complexity. This approach captures both spatial and temporal patterns, with the MP4 compression algorithm serving as an approximation of the minimal description length.

\subsubsection{Multi-scale Matter Distribution}

To capture the spatial organization at different scales, we compute the entropy of matter distribution at various resolutions. For a given state $S$, we define a series of downscaled representations $S^1, S^2, ..., S^n$ and compute the entropy for each:

\begin{equation}
    H_i = -\sum_{x,y} p_i(x,y) \log p_i(x,y)
\end{equation}

where $p_i(x,y)$ is the proportion of matter at position $(x,y)$ in the downscaled representation $S^i$. This provides insights into hierarchical organization, from fine-grained local patterns to large-scale global structures. This metric is computed for the final state of the simulation.

\subsection{IMGEP Implementation}

We employ an Intrinsically Motivated Goal Exploration Process (IMGEP) algorithm, as outlined in Algorithm \ref{alg:imgep}. The IMGEP implementation used here iteratively samples target behavioral descriptors as goals, then searching for Flow-Lenia parameters that lead to these goals (using independent per-dimension Gaussian noise over known parameters that are closest to these goals). This progressively leads to discovering and illuminating Flow-Lenia simulations that are diverse in that space of descriptors (here system-level metrics). Our parameter space includes  parameters related to kernels, growth functions and local rules, along with environmental factors such as the strength of diffusion and mutation rate.

\begin{algorithm}
    \caption{IMGEP for Exploring System-wide Properties in Flow-Lenia}
    \label{alg:imgep}
    \begin{algorithmic}[1]
        \Require Length of the bootstrapping phase $N$
        \State Initialize an empty database of discoveries
        \For{each iteration}
        \If{Number of trials $< N$}
        \State Sample random initial parameters
        \Else
        \State Sample random goal in system-level metrics space
        \State Find closest explored goal and its parameters
        \State Mutate the parameters
        \EndIf
        \State Set up Flow-Lenia with the chosen parameters
        \State Run Flow-Lenia simulation
        \State Compute achieved metrics from simulation results
        \State Add new parameters and achieved metrics to database
        \EndFor
    \end{algorithmic}
\end{algorithm}

To address the computational intensity of long-term simulations, we utilized parallel simulations with reduced grid sizes, exploring a broader range of configurations within computational constraints.

\section{Experiments}

Our experiments aim to uncover a wide range of behaviors, as measured by the metrics that define the IMGEP goal-space. We first examine how different mixing rules affect evolutionary dynamics, which motivates our choice of the negotiation rule for the experiments that follow. We then conduct two IMGEP-based experiments: one investigating Flow-Lenia's ability to exhibit ecosystem-like dynamics, and another focusing on discovering movement patterns of small amounts of matter in environments with obstacles. Finally, leveraging the archive built during the ecosystem experiment, we run a scaling study across six spatial scales and seven time horizons. We analyze how the resulting universes change in terms of reached metrics and local parameter distributions, and we observe larger-scale patterns appearing. Together, these experiments demonstrate the versatility of our method and illustrate how a diversity search on relatively smaller-scale worlds serve as a principled scaffold for designing subsequent, larger experiments. We complement the quantitative results with qualitative analysis to highlight the diverse dynamics that Flow-Lenia can produce, supported by an interactive exploration tool\footnote{\label{footnote1}\url{https://developmentalsystems.org/Flow-Lenia-Universes-Journal/}} that enables examination of the relationship between parameter configurations, reached goals, and videos generated from intermediate simulation steps.

\subsection{Mixing Rules and Evolutionary Activity}
\label{sec:mixing_experiments}

Before exploring Flow-Lenia's parameter space with IMGEP, we investigated how different mixing rules affect evolutionary activity. This comparison informed our choice of the negotiation rule for subsequent experiments.

\subsubsection{Experimental Setup}

We ran simulations on a $512 \times 512$ grid with $3$ channels and $45$ kernels for $5 \times 10^{5}$ time steps. We tested several mixing rules defined in section~\ref{sec:mixing_rules}: weighted average, stochastic selection (from prior Flow-Lenia work), stochastic selection with softmax probabilities, stochastic gene-wise selection, dot-product-based selection, and the negotiation rule introduced in this paper. All simulations used identical initial conditions and identical kernel and growth-function parameters; only the mixing rule varied.

\subsubsection{Results}

\begin{figure}[ht]
    \centering
    \includegraphics[width=0.9\columnwidth]{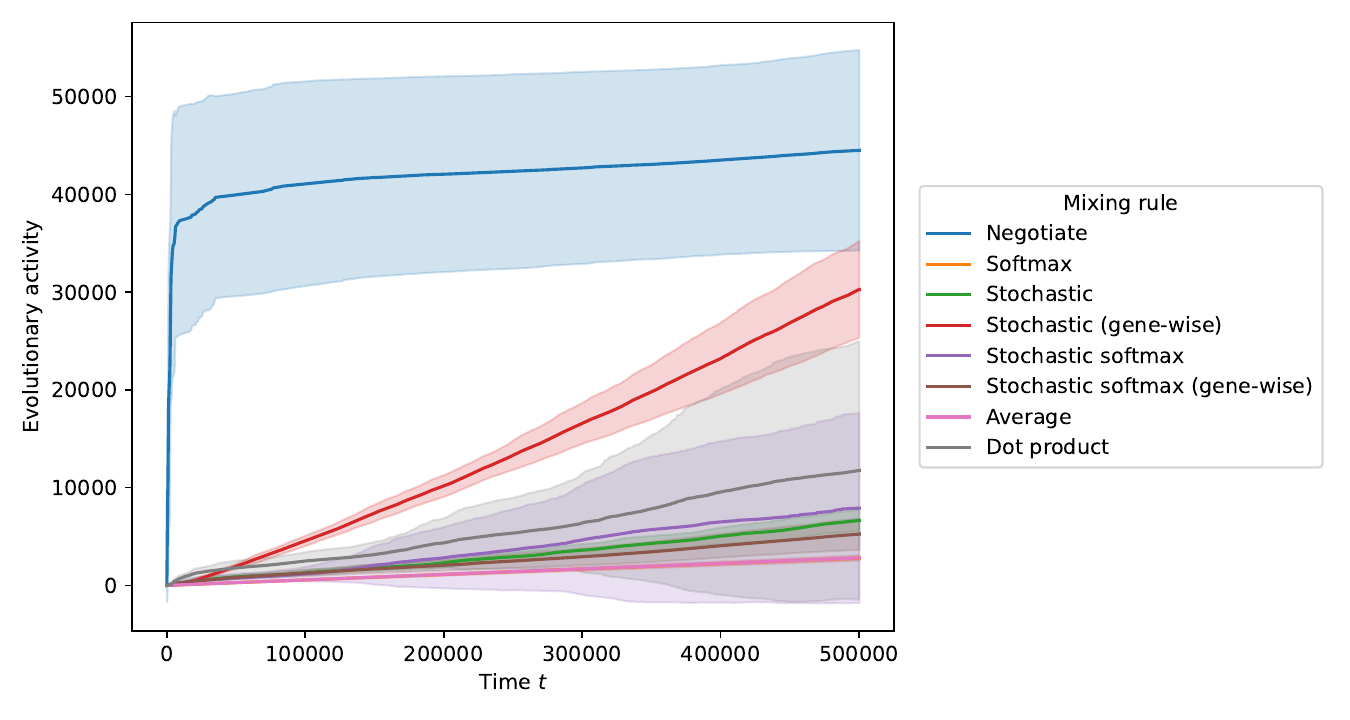}
    \caption{Non-neutral evolutionary activity over time for each mixing rule ($\beta = 100$ for the negotiation rule). Shaded regions show one standard deviation across five seeds; measurements sampled every $250$ simulation steps.}
    \label{fig:ea_over_time}
\end{figure}

Figure~\ref{fig:ea_over_time} reports the non-neutral evolutionary activity over time for each mixing rule. The negotiation rule produced the highest sustained activity, with a rapid initial rise followed by slower but continued growth. Stochastic gene-wise selection came second with near-linear growth. The stochastic softmax and dot-product rules reached high activity but with much larger variance across seeds; the dot-product rule is the only deterministic rule that achieves high activity. The stochastic rule used in previous Flow-Lenia studies~\citep{plantec2023flow} reached lower activity than these alternatives. Averaging rules showed almost no activity: at each pixel, the parameter becomes a continuous blend of all incoming parameters, producing nearly as many distinct genomes as there are pixels, and these genomes do not expand spatially.

\begin{figure}[ht]
    \centering
    \includegraphics[width=0.9\columnwidth]{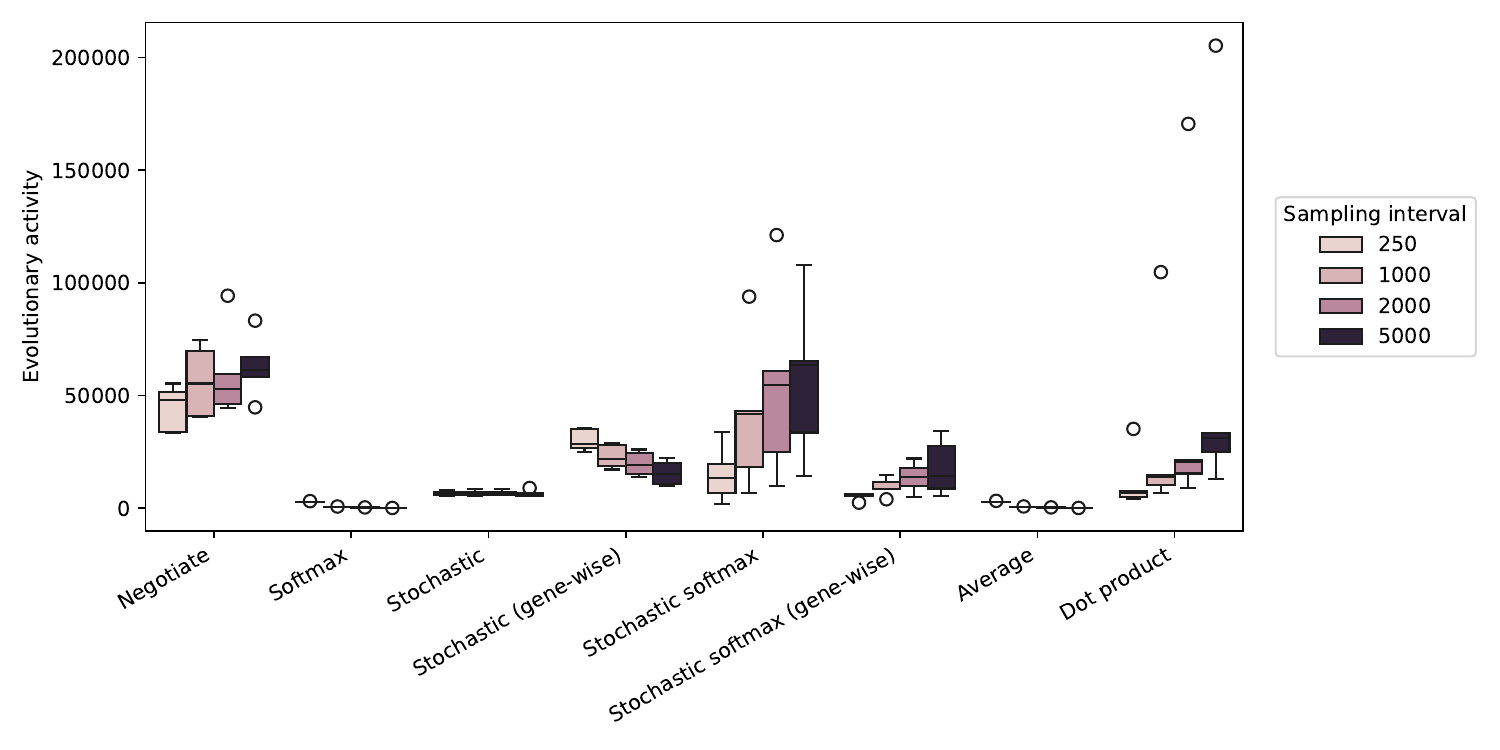}
    \caption{Final evolutionary activity at $t = 5 \times 10^{5}$ for each mixing rule, across sampling intervals of $250$, $1000$, $2000$, and $5000$ simulation steps between measurements.}
    \label{fig:ea_final}
\end{figure}

Figure~\ref{fig:ea_final} shows the final activity values across sampling intervals. The ranking from Figure~\ref{fig:ea_over_time} carries over for every sampling interval we tested.

Larger sampling intervals yield higher per-step activity for every rule, since they capture longer-term population changes rather than short-term fluctuations.

\subsubsection{Discussion}

Mixing rules shape evolutionary dynamics in Flow-Lenia. The rules that produce the highest activities (the negotiation rule, stochastic gene-wise selection, and to a lesser extent the dot-product and stochastic softmax rules) share the property that they preserve parameter identity rather than blending parameters together, which sustains genetic diversity across the grid. Stochasticity helps but is not strictly required: the dot-product rule is deterministic and still reaches high activity in some seeds.

The negotiation rule's advantage stems from its use of affinity information: parameters that achieve high affinity in their local environment are more likely to persist, even when facing competition from larger quantities of matter. This creates selection pressure that favors parameters adapted to their context, driving sustained evolutionary dynamics.

This led us to adopt the negotiation rule for the IMGEP experiments that will follow, for two main reasons. First, it overall produces the highest evolutionary activity. Second, most of the activity increase occurs in a short time frame ($<10^4$ time steps) which matches with the relatively low duration of the small-scale IMGEP runs we will present. Despite the fact that evolutionary activity does not capture every aspect of open-ended evolution (e.g., spatial diversity, temporal complexity, and emergent behaviors require additional metrics), it still provides a useful measure of whether the system supports ongoing adaptive change.

\subsection{Exploring Ecosystem-like Dynamics}
\label{sec:exploring_ecosystems}

\begin{figure}[ht]
    \centering
    \includegraphics[width=1.0\columnwidth]{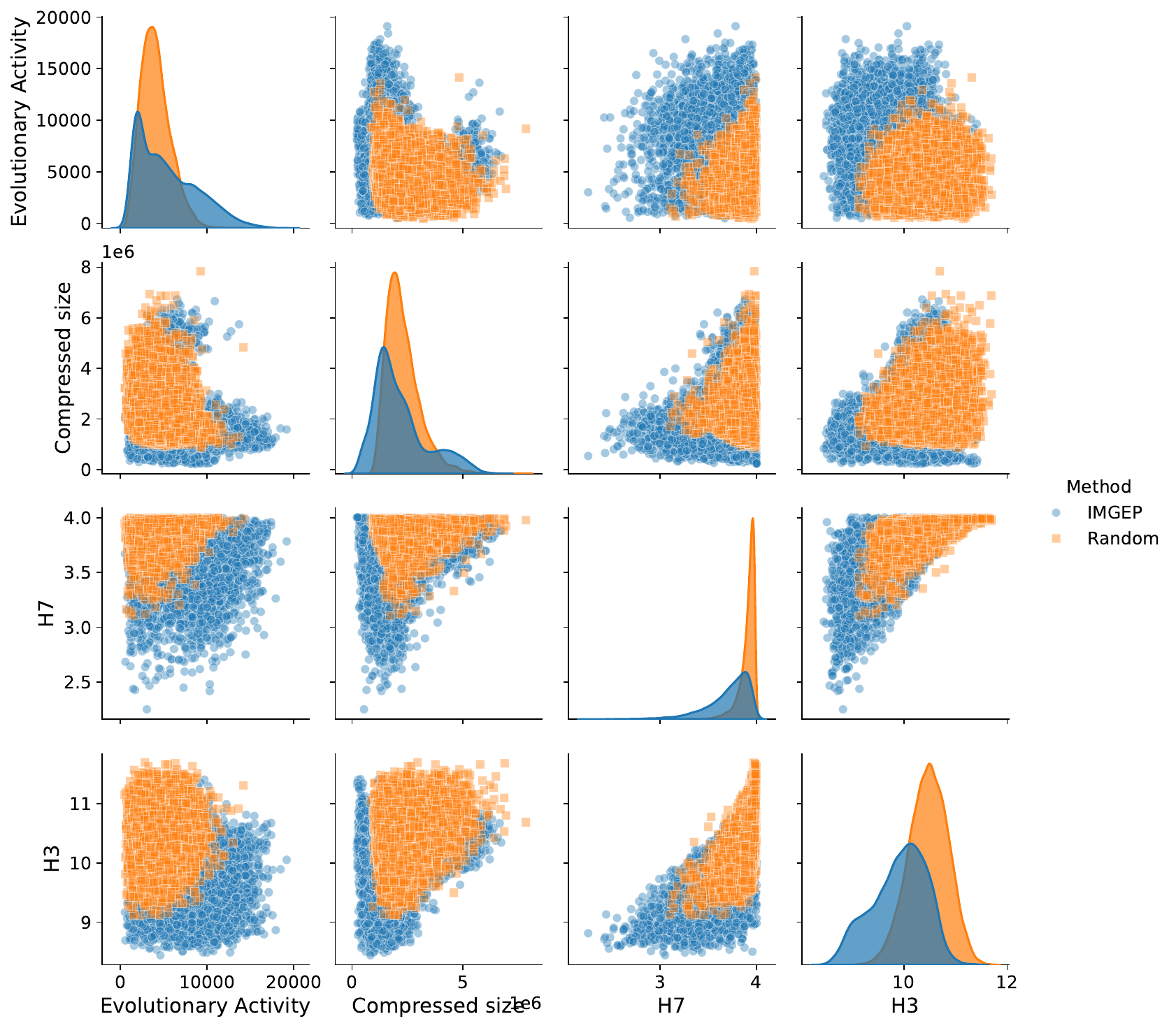}
    \caption{Goal-space coverage comparison between IMGEP and random exploration, showing that IMGEP consistently illuminates more of the goal-space than random search.}
    \label{fig:imgep_exploration}
\end{figure}

The first IMGEP experiment uses the metrics described in Section \ref{metrics} to discover diverse Flow-Lenia universes. We use a 256$\times$256 grid with 3 channels and 9 kernels, with simulations running for 10,000 time steps.
Figure \ref{fig:imgep_exploration} shows the goals reached with our method compared to random search (i.e., uniform sampling in parameter space). The difference is most evident in evolutionary activity and multi-scale entropy metrics.

To evaluate the effectiveness of our method, we compute the average pairwise distance in the goal-space, finding that IMGEP achieves a distance of 0.874, significantly higher than the 0.476 obtained through random search. Additionally, we compute a binning-based coverage by discretizing each goal-space dimension into five bins and counting the number of non-empty bins. IMGEP identifies 576 non-empty bins, in contrast to only 205 discovered via random sampling.

\begin{figure}[tbp]
    \centering
    \begin{subfigure}[c]{0.38\columnwidth}
        \centering
        \footnotesize
        \setlength{\tabcolsep}{3pt}
        \renewcommand{\arraystretch}{1.1}
        \begin{tabular}{l|cc}
            \hline
            \textbf{Metric} & \textbf{IMGEP} & \textbf{Random} \\
            \hline
            Avg Pair. Dist. & \textbf{8.74e-01} & 4.76e-01 \\
            Coverage        & \textbf{576}      & 205      \\
            EA              & \textbf{3.44e+03} & 1.70e+03 \\
            MP4             & \textbf{1.20e+06} & 7.60e+05 \\
            H7              & \textbf{2.26e-01} & 8.62e-02 \\
            H6              & \textbf{4.10e-01} & 1.66e-01 \\
            H5              & \textbf{5.43e-01} & 2.54e-01 \\
            H4              & \textbf{5.67e-01} & 3.44e-01 \\
            H3              & \textbf{5.19e-01} & 3.79e-01 \\
            \hline
        \end{tabular}
        \caption{Exploration performance. Higher values indicate broader exploration of evolutionary dynamics.}
        \label{tab:imgep_performance}
    \end{subfigure}%
    \hfill
    \begin{subfigure}[c]{0.60\columnwidth}
        \centering
        \includegraphics[width=\linewidth]{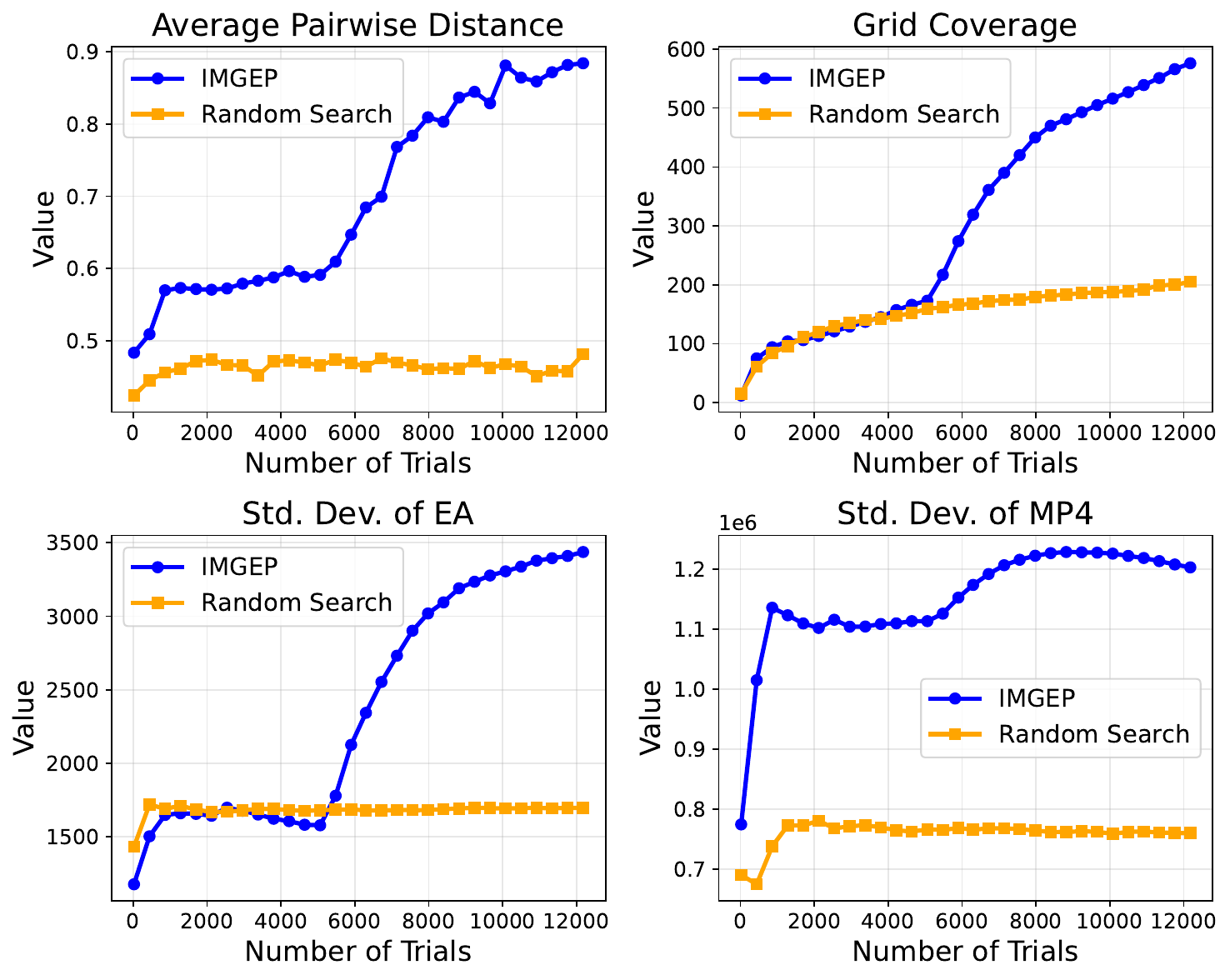}
        \caption{Evolution of exploration metrics over time. The consistent divergence between IMGEP (blue) and random search (orange) indicates IMGEP's sustained ability to discover new evolutionary regimes.}
        \label{fig:imgep_overtime}
    \end{subfigure}
    \caption{Comparison of the metric coverage between IMGEP and random exploration in the ecosystem-dynamics experiment.}
    \label{fig:exploration_performance}
\end{figure}

Sharp transitions in the exploration metrics over time (Figure \ref{fig:imgep_overtime}) show that IMGEP found a subset of parameters that strongly increase diversity in goal-space (around trial 6000). The ongoing increase in coverage suggests that many areas of the metric space have yet to be explored.

\begin{figure}[tbp]
    \centering
    \begin{minipage}[c]{0.51\columnwidth}
        \centering
        \begin{tabular}{@{\hskip 2pt}c@{\hskip 2pt}c@{\hskip 2pt}c@{\hskip 2pt}c@{\hskip 2pt}}
            \includegraphics[width=0.21\linewidth]{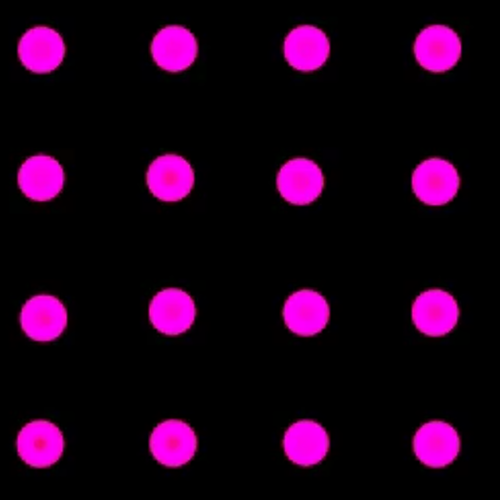} &
            \includegraphics[width=0.21\linewidth]{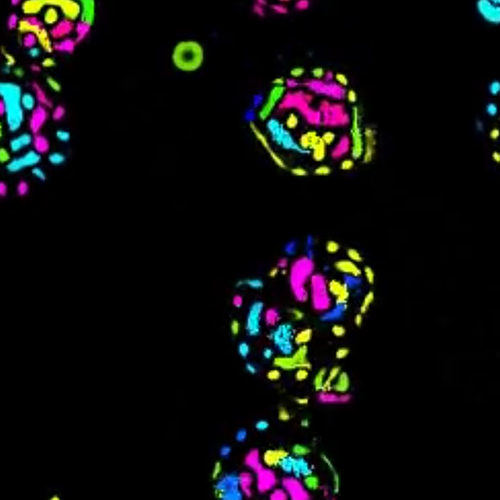} &
            \includegraphics[width=0.21\linewidth]{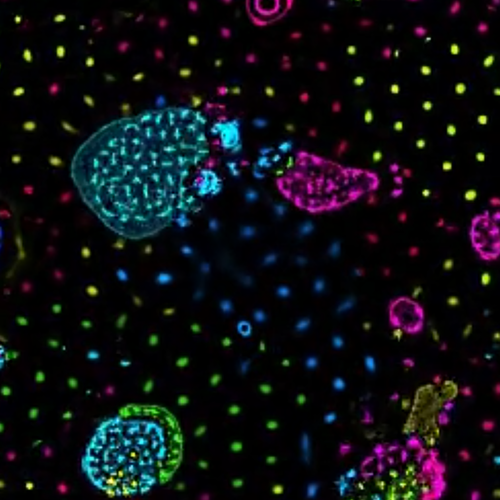} &
            \includegraphics[width=0.21\linewidth]{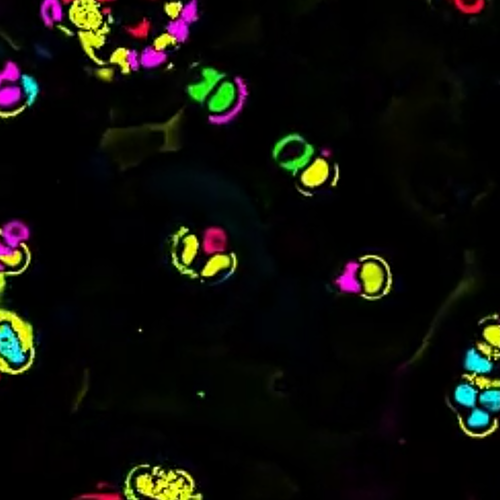}   \\
            \includegraphics[width=0.21\linewidth]{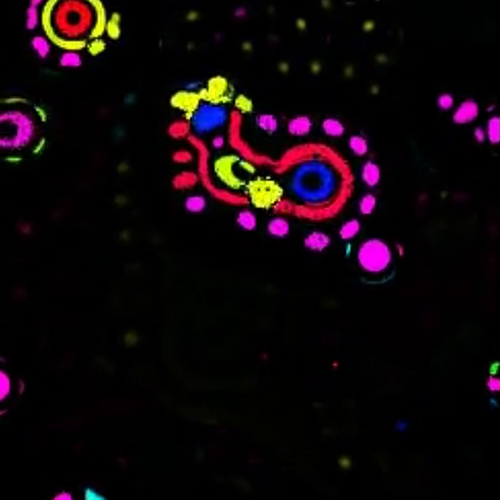} &
            \includegraphics[width=0.21\linewidth]{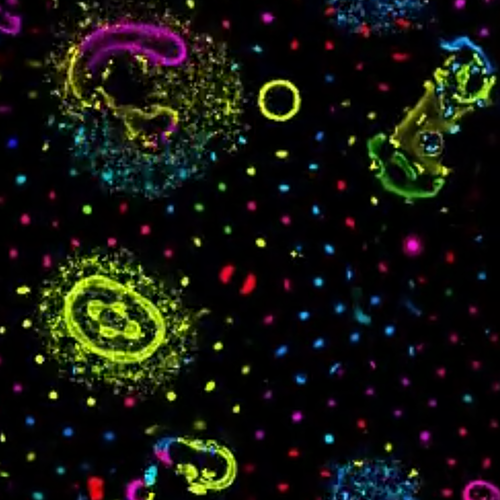} &
            \includegraphics[width=0.21\linewidth]{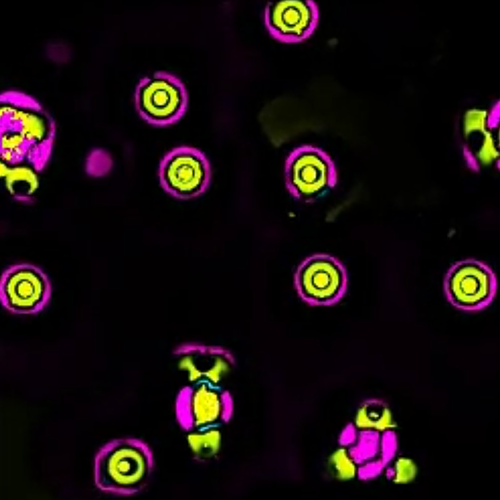} &
            \includegraphics[width=0.21\linewidth]{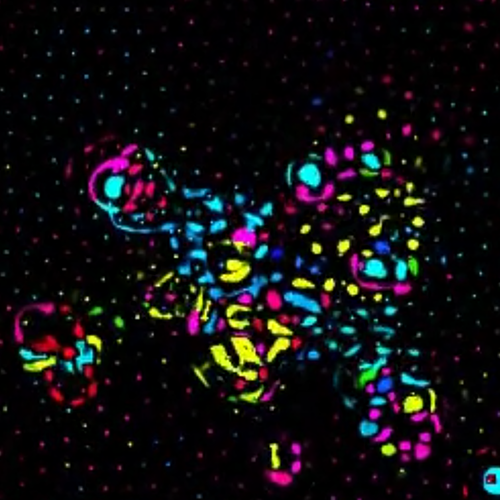}   \\
        \end{tabular}
    \end{minipage}%
    \hfill
    \begin{minipage}[c]{0.48\columnwidth}
        \centering
        \begin{tabular}{@{\hskip 2pt}c@{\hskip 2pt}c@{\hskip 2pt}c@{\hskip 2pt}}
            \includegraphics[width=0.32\linewidth]{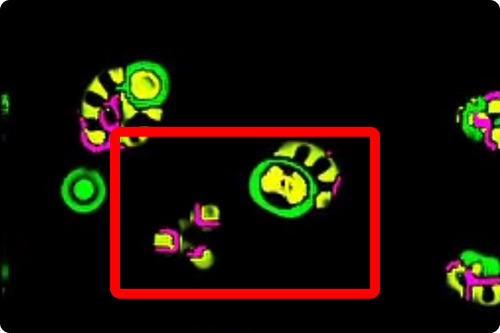} &
            \includegraphics[width=0.32\linewidth]{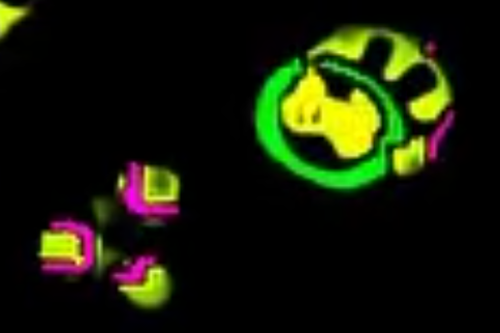} &
            \includegraphics[width=0.32\linewidth]{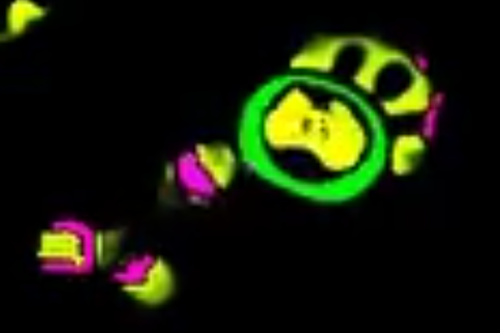}   \\
            \includegraphics[width=0.32\linewidth]{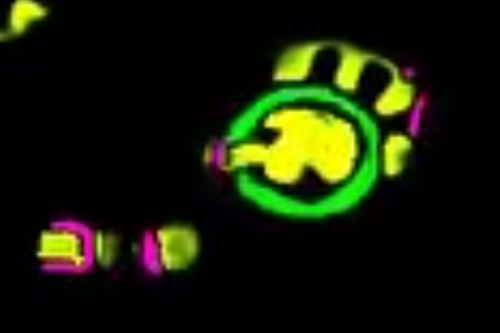} &
            \includegraphics[width=0.32\linewidth]{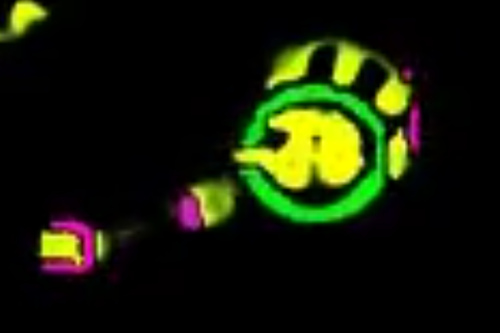} &
            \includegraphics[width=0.32\linewidth]{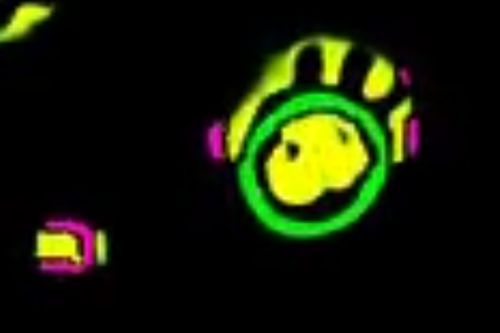}   \\
        \end{tabular}
    \end{minipage}
    \caption{Ecosystem-dynamics experiment: (left) discovered diversity of self-organized patterns; (right) a temporal sequence illustrating feeding-like behavior.}
    \label{fig:ecosystem_qualitative}
\end{figure}

A closer examination of the videos in the discovery archive (sample frames shown in Figure \ref{fig:ecosystem_qualitative}) reveals a striking diversity of self-organizing behaviors, both spatially and temporally. Some universes show the spread of a dominant local parameter that gradually takes over all matter, leading to convergence to an orderly state. Others exhibit chaotic dynamics, where matter continuously merges, separates and mutates, with older patterns fading as new ones emerge. In some cases, large, cohesive structures form from multiple localized parameters types, while others resemble dense clusters akin to bacterial colonies. We also observe behaviors reminiscent of feeding, where larger patterns composed of several localized parameters consume smaller ones (see Figure \ref{fig:ecosystem_qualitative}, right).

\subsection{Exploring Matter Movement}

\begin{figure}[bp]
    \centering
    \begin{minipage}[c]{0.18\columnwidth}
        \centering
        \begin{tabular}{@{}c@{}}
            \includegraphics[width=0.95\linewidth]{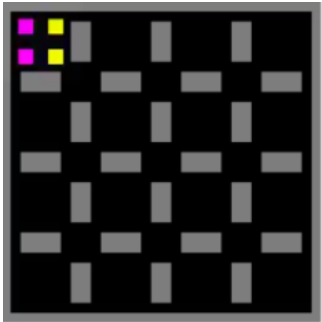} \\
            \footnotesize Initial state \\[4pt]
            \includegraphics[width=0.95\linewidth]{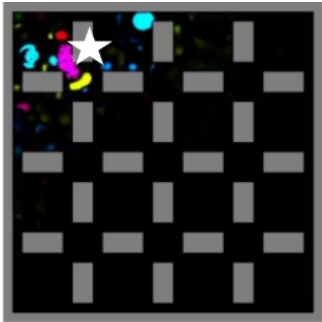} \\
            \footnotesize Final state \\
        \end{tabular}
    \end{minipage}%
    \hfill
    \begin{minipage}[c]{0.80\columnwidth}
        \centering
        \renewcommand{\tabcolsep}{1pt}
        \begin{tabular}{cccc}
            \includegraphics[width=0.22\linewidth]{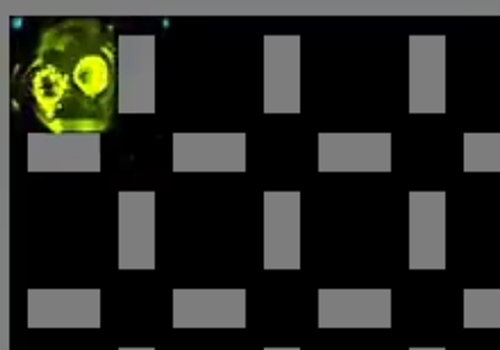}        &
            \includegraphics[width=0.22\linewidth]{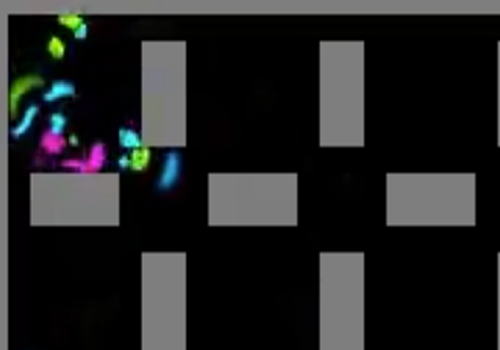}         &
            \includegraphics[width=0.22\linewidth]{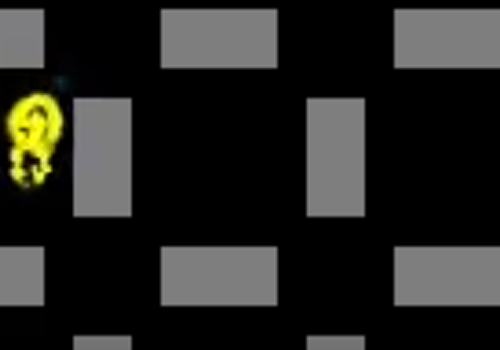} &
            \includegraphics[width=0.22\linewidth]{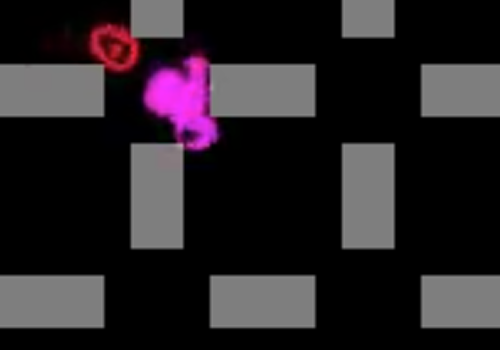}  \\
            \includegraphics[width=0.22\linewidth]{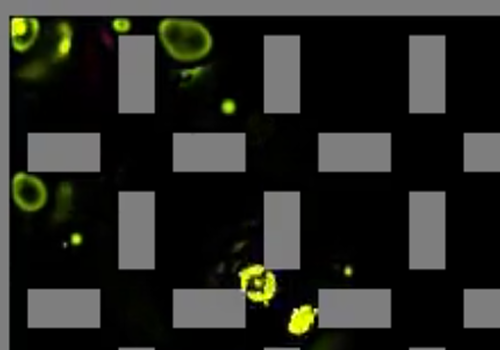}        &
            \includegraphics[width=0.22\linewidth]{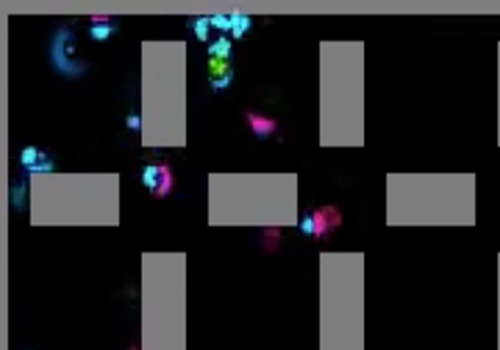}         &
            \includegraphics[width=0.22\linewidth]{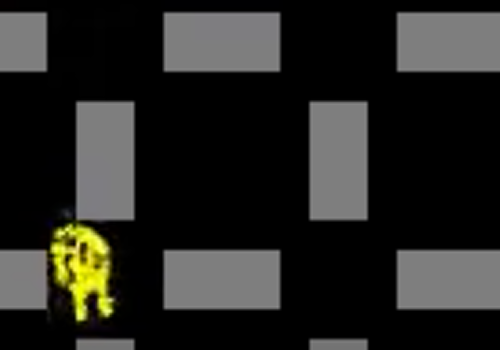} &
            \includegraphics[width=0.22\linewidth]{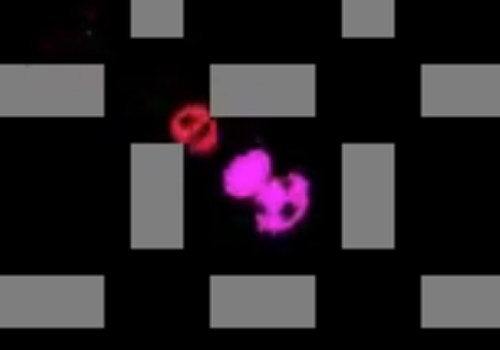}  \\
            \includegraphics[width=0.22\linewidth]{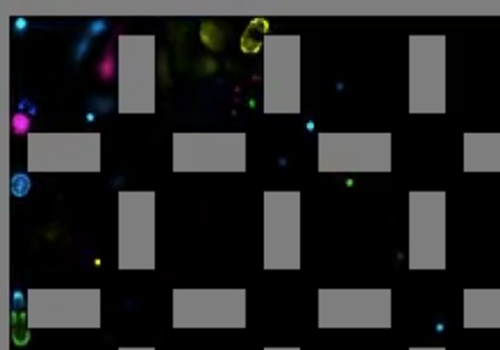}        &
            \includegraphics[width=0.22\linewidth]{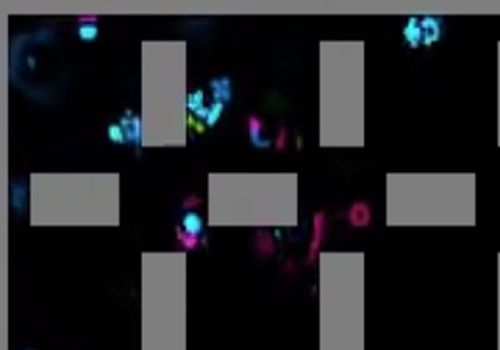}         &
            \includegraphics[width=0.22\linewidth]{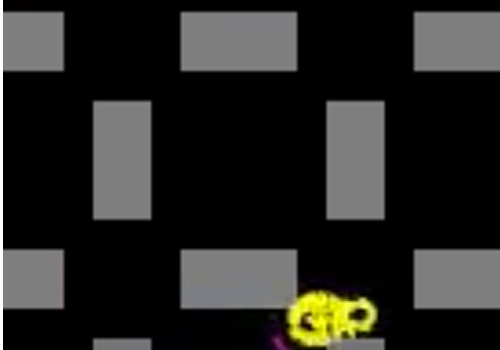} &
            \includegraphics[width=0.22\linewidth]{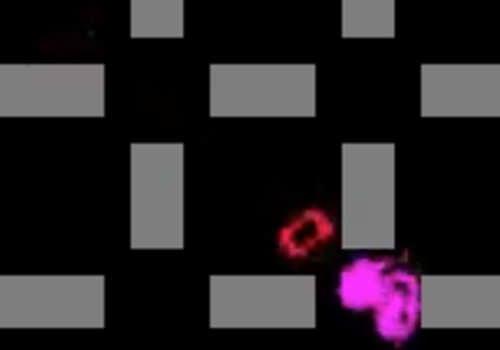}  \\
            a) & b) & c) & d) \\
        \end{tabular}
    \end{minipage}
    \caption{Matter-movement experiment. \textbf{Left:} Initial and final states of a representative run: matter starts concentrated in one corner of a walled grid; the white star marks the center of mass (used as the IMGEP goal-space). \textbf{Right:} A gallery of different movement regimes discovered through IMGEP exploration, each column showing a temporal sequence: a)~allopatric speciation, b)~initial pattern splitting into agile movers, c)~pattern turning in a corridor, d)~direct movement.}
    \label{fig:matter_movement}
\end{figure}

The second set of experiments focuses on discovering diverse patterns of matter movement within environments that include obstacles. We design an environment composed of a grid of walls with passages and place all the initial matter in one corner of the grid (Figure \ref{fig:matter_movement}, left). We explore such environments with IMGEP through a simple goal-space metric: the center of mass of all matter at the end of the simulation (Figure \ref{fig:matter_movement}, right).

\begin{figure}[t]
    \centering
    \begin{tabular}{cc}
        \includegraphics[width=0.45\columnwidth]{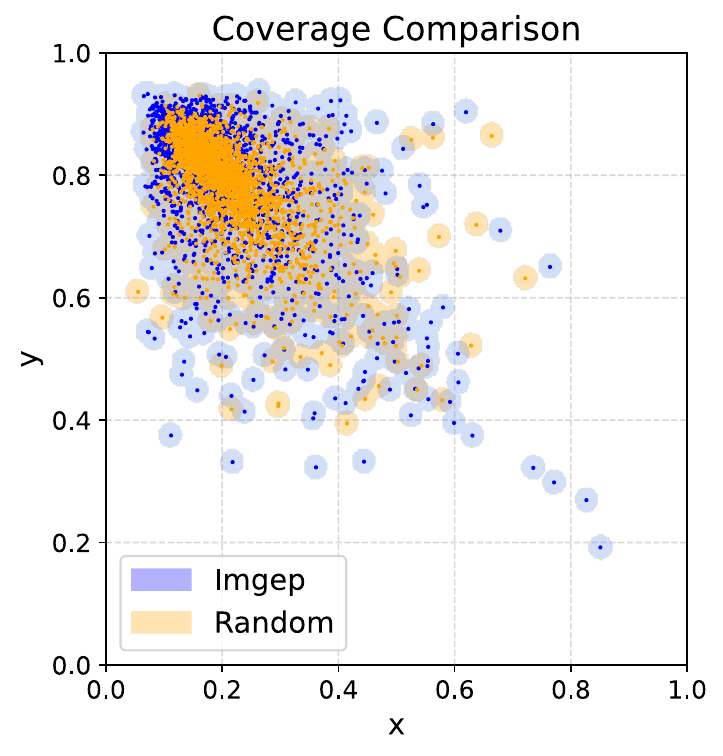} &
        \includegraphics[width=0.45\columnwidth]{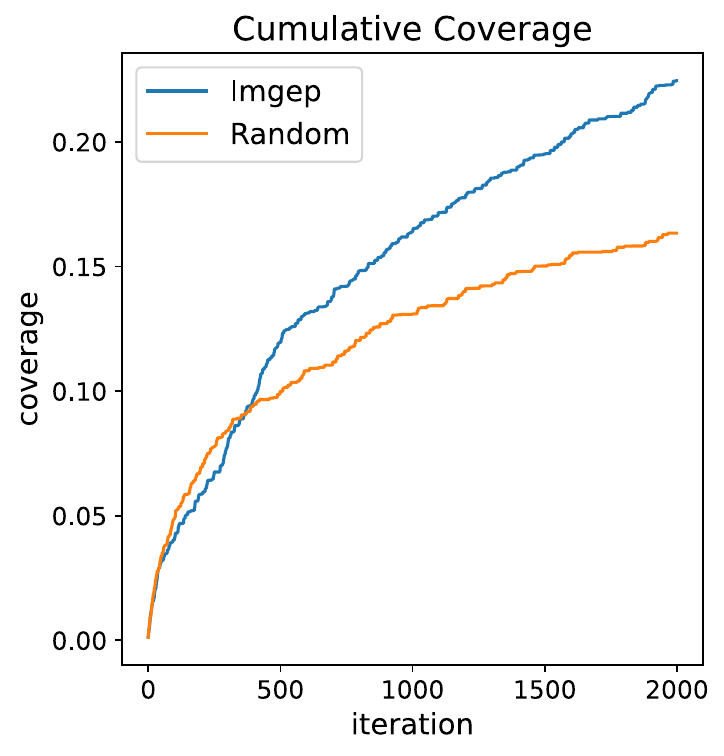} \\
    \end{tabular}
    \caption{Comparison of reached goals \textbf{(left)} and cumulative coverage curves \textbf{(right)} between IMGEP and random search in the matter distribution experiment.}
    \label{fig:matter_dist_result}
\end{figure}

We run IMGEP for 2000 iterations and compare its performance to random exploration. As shown in Figure \ref{fig:matter_dist_result}, IMGEP achieves broader coverage in the goal-space and reaches points corresponding to matter that has traveled farther from its initial location. This indicates that IMGEP can identify simulation parameters that enable diverse movement behaviors.

Qualitative analysis reveals a rich variety of movement dynamics. We observe both directed and chaotic motion, as well as large patterns fragmenting into smaller, faster-moving entities capable of navigating through narrow corridors. Other cases show dense clusters of matter that gradually diffuse across the environment, with subsequent mutations giving rise to multiple local parameters --- reminiscent of allopatric speciation (see Figure \ref{fig:matter_movement}).

\subsection{Scaling Discovered Ecosystems in Space and Time}
\label{sec:scaling}

Returning to the ecosystem dynamics experiment of Section~\ref{sec:exploring_ecosystems}, we now investigate how the discovered universes change when they are scaled in space and time. For this aim, we leverage the previously generated IMGEP archive as a springboard: the already illuminated diverse universes provide a curated starting set from which to launch a focused study of how scaling affects goal-space metrics, the local-parameter composition of the ecosystems, and their qualitative behavior. Compared to rerunning exploration from scratch at larger and more expensive scales, this method enables to discover large-scale ecosystemic dynamics at much lower computational cost.

\subsubsection{Experimental Setup}

\begin{figure}[ht]
    \centering
    \includegraphics[width=1.0\columnwidth]{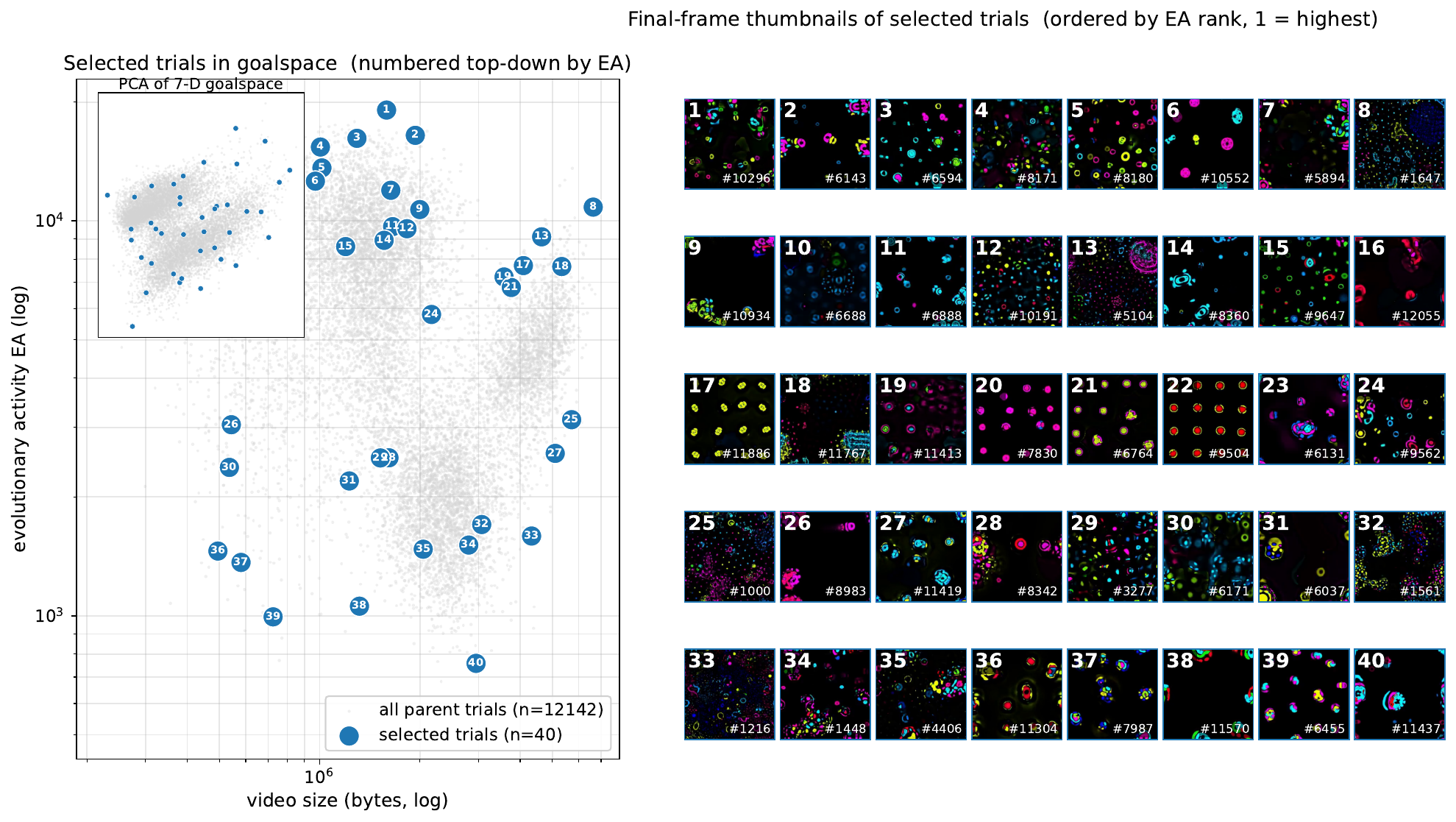}
    \caption{Diverse trials sampled by farthest-point sampling in the z-scored goal-space from the IMGEP archive of Section~\ref{sec:exploring_ecosystems}, shown in the EA $\times$ MP4 video size plane, together with the final-frame thumbnail of each selected trial. These trials are used in the scaling study.}
    \label{fig:scaling_selection}
\end{figure}

We select 40 diverse trials from the archive of Section~\ref{sec:exploring_ecosystems} using farthest-point sampling on the z-scored seven-dimensional goal-space. Starting from a single randomly chosen trial, the algorithm iteratively adds the archive entry whose minimum distance to the current selection is largest, producing a subset that spans the goal-space rather than oversampling any one regime. Figure \ref{fig:scaling_selection} visualizes the selected trials in the archive goal-space together with thumbnails of their final states.

Each selected trial is rerun on the Cartesian product of six spatial scales (grid sides multiplied by $k$ for $k \in \{1,\dots,6\}$, so the area scales as $k^2$) and seven horizons (5{,}000, 10{,}000, 20{,}000, 40{,}000, 80{,}000, 160{,}000, and 320{,}000 time steps), yielding $40 \times 6 \times 7 = 1680$ simulations. The mixing rule, kernel and growth-function parameters, and per-area mutation rate are held fixed across scales; only the grid size and the horizon vary. Initial conditions at scale $k$ are obtained by tiling the base $256\times256$ initialization, so that each scaled run starts from the same local statistics as its base-scale counterpart.

Several of the metrics introduced in Section \ref{metrics} grow mechanically with grid area or horizon, which makes raw cross-scale comparisons misleading. We therefore complement raw evolutionary activity (EA) and the MP4 metric with normalized variants. Normalized EA divides total EA by the product of mean grid mass and the horizon length, expressing activity on a per-unit-mass per-time-step basis. Normalized MP4 divides total size by the number of grid cells and the horizon length, expressing per-cell, per-time-step information content.
\subsubsection{Results}
\begin{figure}[tbp]
    \centering
    \includegraphics[width=1.0\columnwidth]{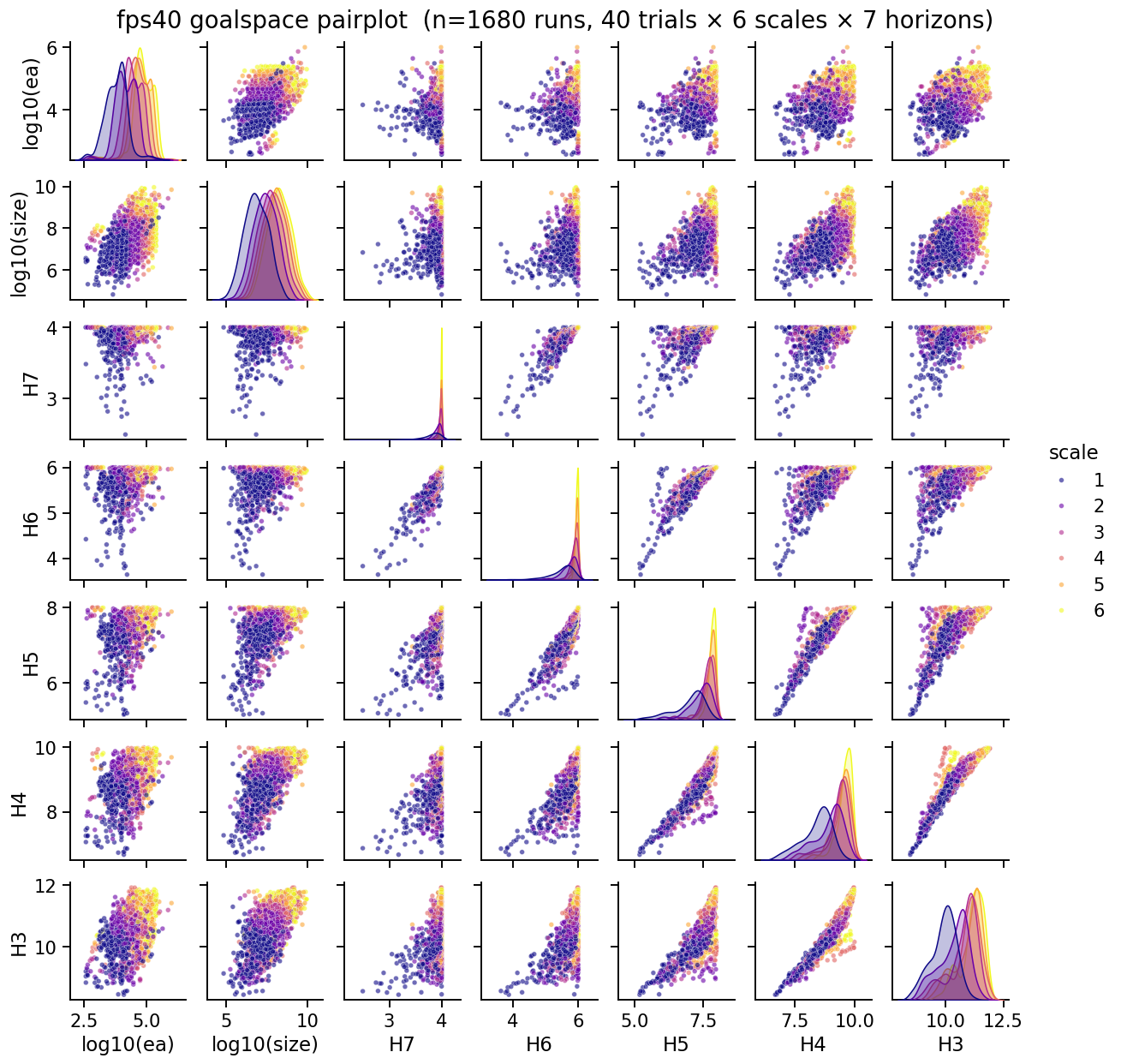}
    \caption{Full goal-space pairplot across all 1680 scaled simulations, with marker color encoding spatial scale.}
    \label{fig:scaling_pairplot_full}
\end{figure}

\begin{figure}[tbp]
    \centering
    \includegraphics[width=1.0\columnwidth]{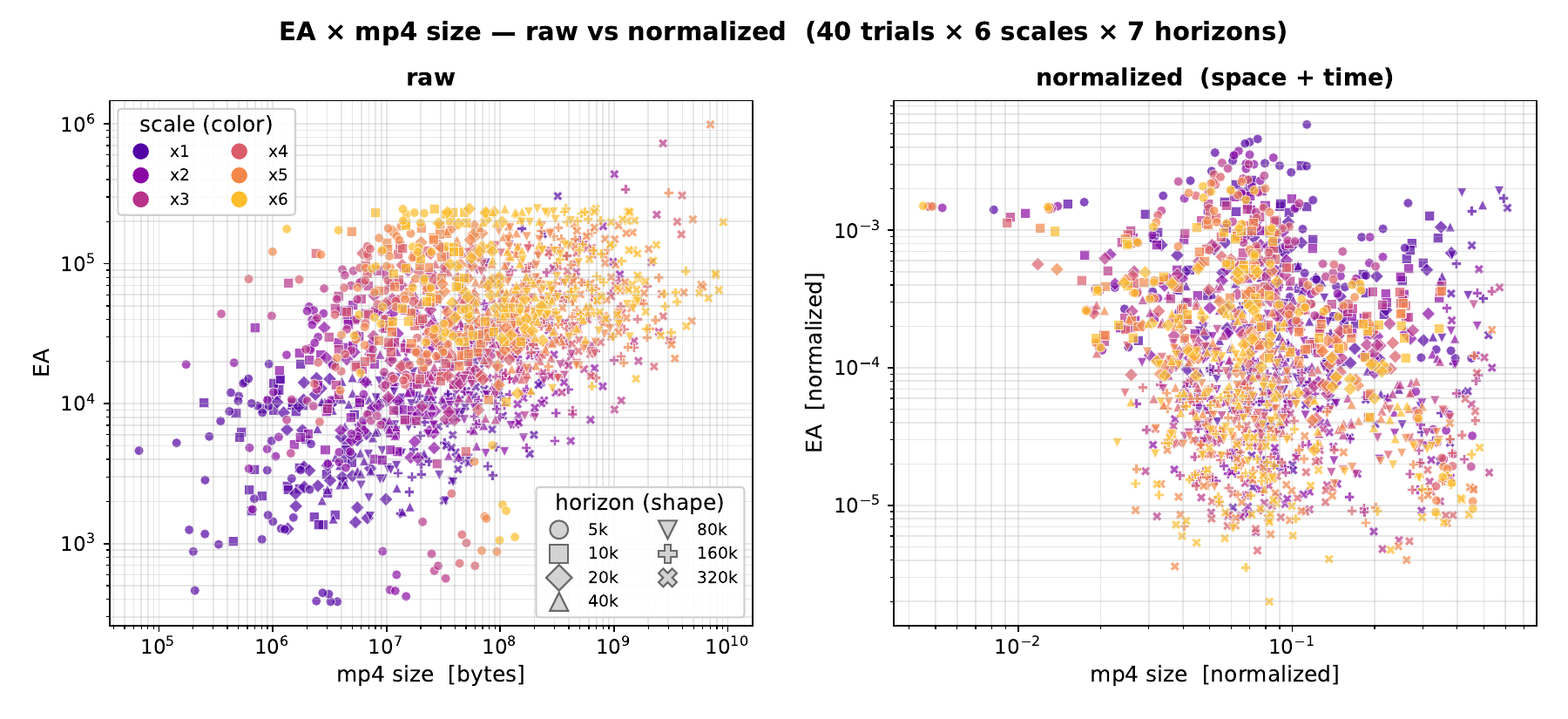}
    \caption{A closer look at raw and normalized evolutionary activity and MP4 size metrics, for individual trials across spatial and temporal scales.}
    \label{fig:scaling_ea_vs_size}
\end{figure}

\begin{figure}[tbp]
    \centering
    \includegraphics[width=0.9\columnwidth]{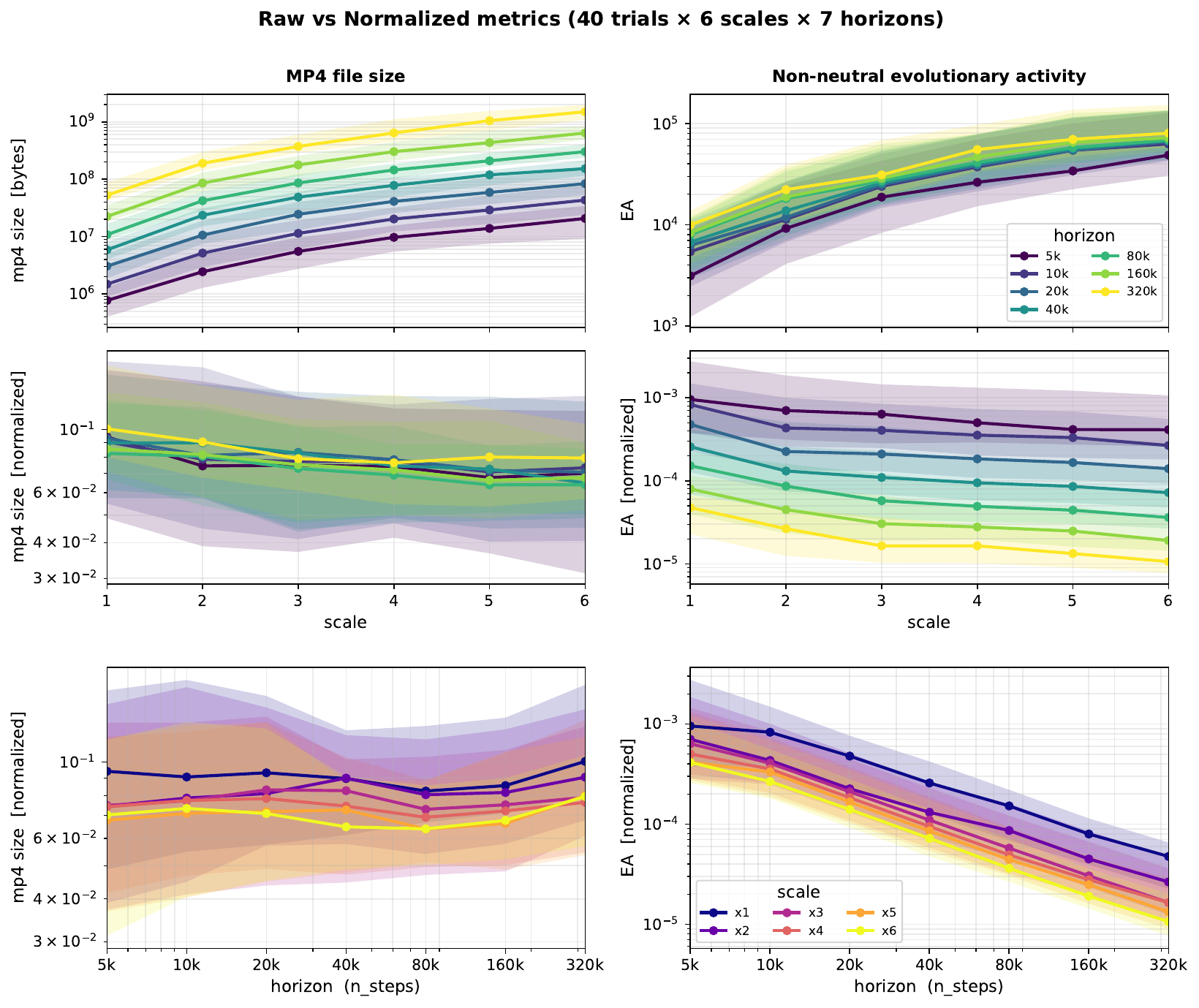}
    \caption{Combined view of raw and normalized longitudinal metrics across spatial scale and horizon. The plots show averaged metrics for all trials across specific temporal and spatial scalesTop row: raw MP4 and EA against scale, colored by horizon. Middle row: the normalized counterparts against scale, same coloring. Bottom row: the normalized counterparts against horizon, colored by scale.}
    \label{fig:scaling_combined}
\end{figure}

Figure \ref{fig:scaling_pairplot_full} reports the raw goal-space metrics across all scaled simulations. Figures \ref{fig:scaling_ea_vs_size} and \ref{fig:scaling_combined} zoom in on EA and MP4, in both raw and normalized form. The raw metrics behave as expected: total EA and total MP4 size grow monotonically with both scale and horizon. Normalization generally inverts this trend — both EA and MP4 decrease as the space and time scales grow — so larger universes carry more total complexity but less complexity per cell and per step.

The EA trend with respect to the horizon (shown in  Figure \ref{fig:scaling_combined}, right column) matches the pattern seen in the mixing rule experiments of Section \ref{sec:mixing_experiments}, Figure \ref{fig:ea_over_time}: EA rises rapidly early in a simulation and much more slowly later on. Qualitatively, our runs tend to pass through three stages. First, matter moves quickly and local parameters change often, producing the rapid color changes visible in our rendering. The universe then organizes into larger, more stable patterns. Finally, for sufficiently long horizons, those macro patterns break apart and the resulting smaller chunks move rapidly and chaotically again. This trajectory accounts for the non-monotonic MP4 horizon dependence in Figure \ref{fig:scaling_combined}, bottom-left: stable middle-stage patterns compress efficiently, while the chaotic early and late stages compress poorly.
Returning to the pairplot in Figure \ref{fig:scaling_pairplot_full}, the coarser entropy estimators — particularly H4 and H8, computed over 4×4 and 8×8 blocks — collapse to nearly identical values across simulations, and the effect becomes more pronounced as grid size grows. In other words, at these spatial scales the strongly coarse-grained metric loses its ability to discriminate between runs.
 
Taken together, the goal-space observations and the qualitative analyses of all experiments give concrete feedback on how to design subsequent experiments. This iterative loop is, we believe, the central strength of our approach: the user defines a set of metrics intended as proxies for interesting, complex phenomena, launches a large-scale diversity search, and then inspects the results with the visualization tool to inform the design of the next experiment.

\begin{figure}[!h]
    \centering
    \begin{subfigure}[c]{0.46\columnwidth}
        \centering
        \small
        \setlength{\tabcolsep}{4pt}
        \begin{tabular}{l|cc}
            \hline
            \textbf{Metric} & \textbf{x1 ($256{\times}256$)} & \textbf{x6 ($1536{\times}1536$)} \\
            \hline
            Unique $P$      & 46.5 $\pm$ 38.4                & 181.5 $\pm$ 163.0                \\
            Mean $P$        & $-10.23 \pm 5.67$              & $-6.23 \pm 3.77$                 \\
            Std $P$         & $22.26 \pm 24.09$              & $12.42 \pm 13.21$                \\
            \hline

    \end{tabular}
        \caption{Aggregated parameter-vector statistics (mean $\pm$ std).}
        \label{tab:scaling_local_params}
    \end{subfigure}%
    \hfill
    \begin{subfigure}[c]{0.52\columnwidth}
        \centering
        \includegraphics[width=\linewidth]{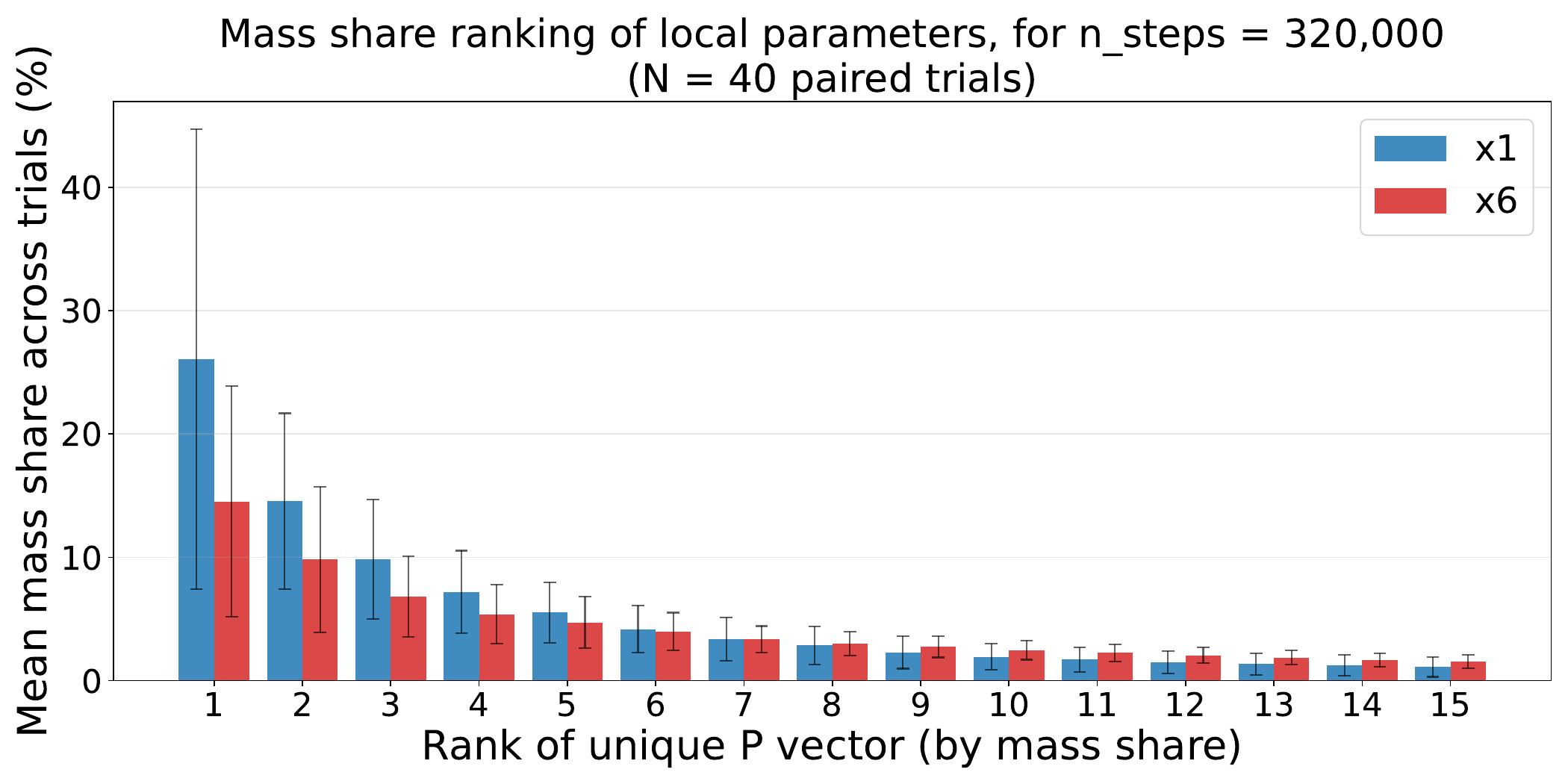}
        \caption{Mass-share distribution across local parameters.}
        \label{fig:scaling_mass_share}
    \end{subfigure}
    \caption{Local-parameter structure at the longest horizon (320{,}000 time steps), comparing the smallest (x1, $256{\times}256$) and largest (x6, $1536{\times}1536$) spatial scales over the 40 sampled trials: (a) aggregated parameter-vector statistics, (b) mass-share distribution across local parameters. Larger grids host more distinct local parameters with smaller magnitudes; mass concentrates in a few dominant rules at x1 and spreads more evenly across many rules at x6.}
    \label{fig:scaling_local_params}
\end{figure}

Beyond aggregate metrics, scale also changes the composition of the ecosystems' local parameters (henceforth \textit{local parameters}). We summarize the population of local parameters present at the end of each run by their count, the magnitude and variance of their parameter vectors, and the share of total mass each rule controls. To isolate the effect of scale from horizon, we restrict this analysis to the longest horizon (320{,}000 time steps) and compare only the extremes of the scaling range, x1 and x6. Table \ref{tab:scaling_local_params} reports parameter-vector statistics aggregated across the 40 trials at these two scales, and Figure \ref{fig:scaling_mass_share} visualizes the corresponding mass-share distribution. At x1, runs typically end with a small number of high-magnitude, high-variance local parameters, and a few rules dominate the mass budget. At x6, runs end with several times as many distinct local parameters, each with smaller magnitude and variance, and the mass distribution across rules is markedly more even.

\begin{figure}[tbp]
    \centering
    \setlength{\tabcolsep}{2pt}
    \begin{tabular}{cc}
        \textbf{x1} ($256{\times}256$) & \textbf{x6} ($1536{\times}1536$) \\[2pt]
        \includegraphics[width=0.38\columnwidth]{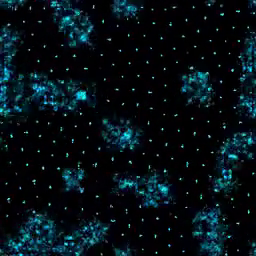}  & \includegraphics[width=0.38\columnwidth]{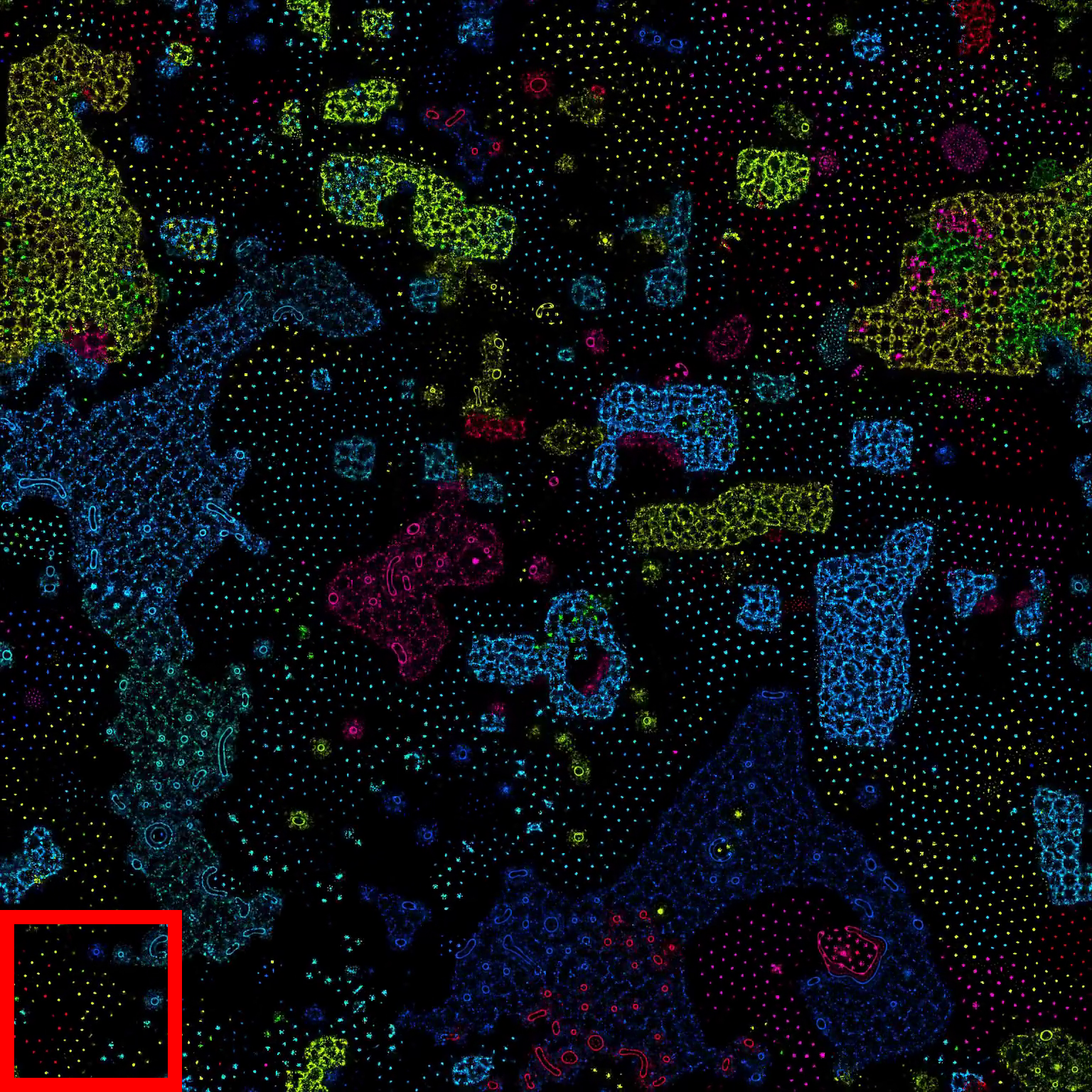}  \\[4pt]
        \includegraphics[width=0.38\columnwidth]{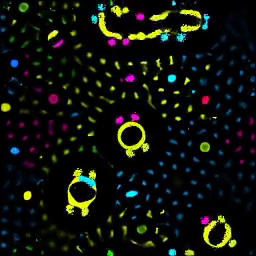}  & \includegraphics[width=0.38\columnwidth]{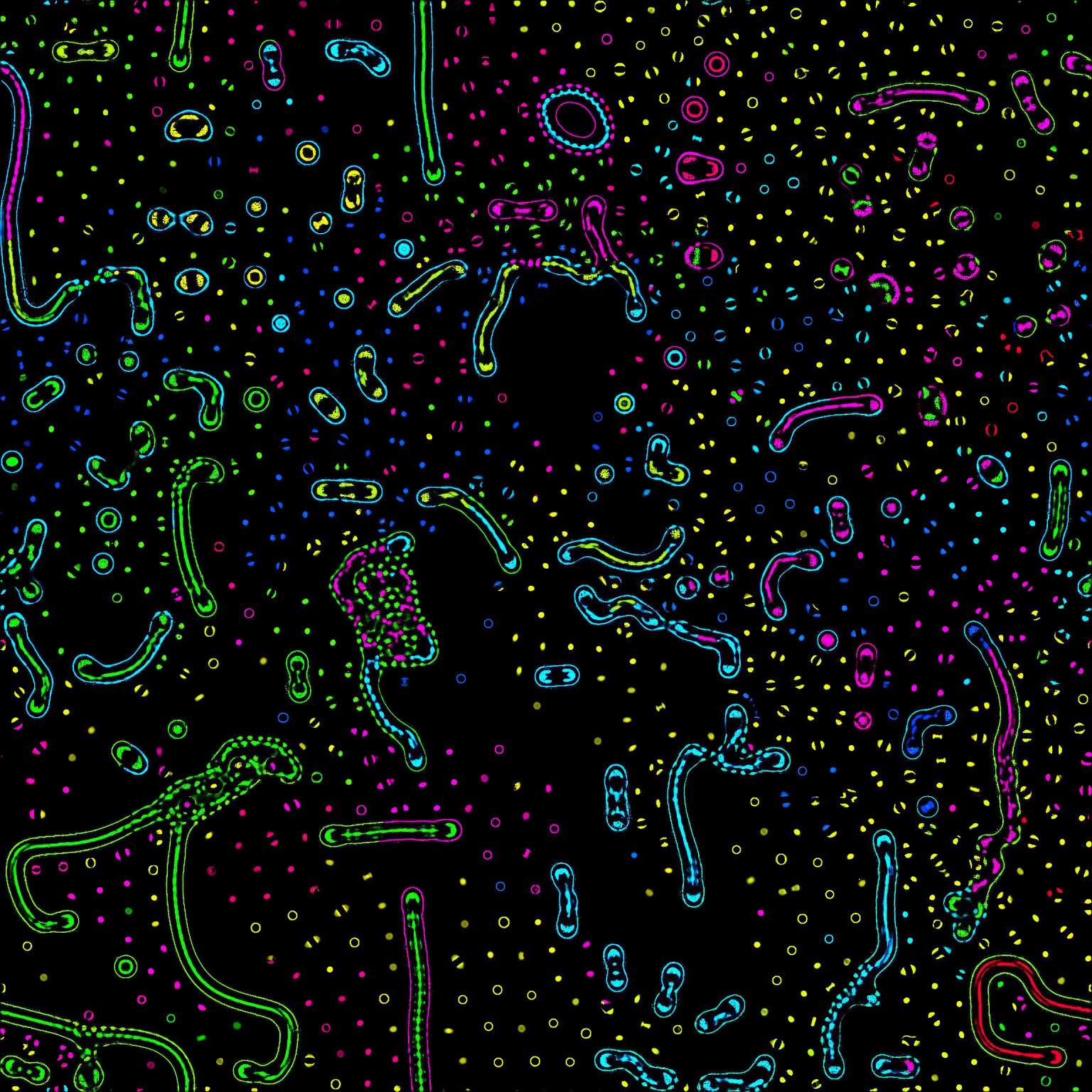}  \\[4pt]
        \includegraphics[width=0.38\columnwidth]{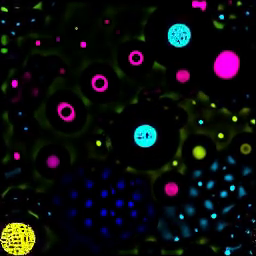} & \includegraphics[width=0.38\columnwidth]{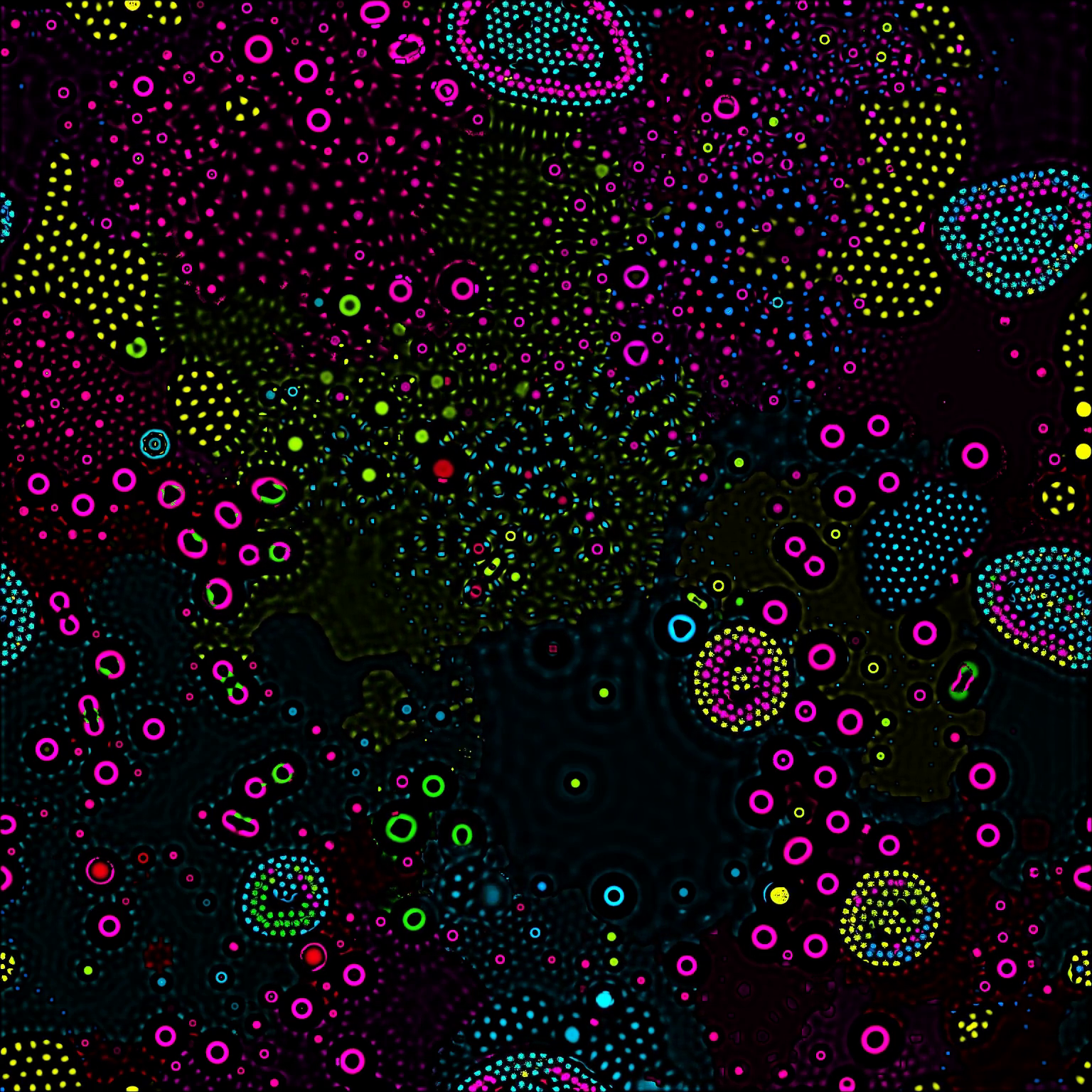} \\
    \end{tabular}
    \caption{A comparison of final states resulting from identical initial local parameters simulated on small ($256{\times}256$) and large ($1536{\times}1536$) grids, for the longest time horizon ($320000$). The red square visualizes the size of the small universes with respect to the large ones. Large universes produce macro-patterns that appear qualitatively more diverse in shape, size, and composition, sometimes being larger than whole small grids.}
    \label{fig:scaling_qualitative}
\end{figure}

The most notable result is the appearance of macro-scale organization. For the largest spatial scale, $1536\times1536$,
several trials produce coherent patterns larger than entire base-sized grids (see Figure \ref{fig:scaling_qualitative} for examples). Although not shown in figures, we note that for largest grid sizes, even the longest horizon (320{,}000 steps) may be insufficient to reach an apparently stationary regime that base-scale runs seem to reach in 10{,}000 to 40{,}000 steps. The macro-patterns observed at $k=6$ may therefore be transient regimes that would resolve into different long-term behavior given still longer simulation.

\section{Discussion and Conclusion}

We adapted IMGEP, a diversity search algorithm, to systematically uncover collective dynamics of complex systems, and applied it to Flow-Lenia to discover ecosystem-like phenomena. While previous applications of diversity search have focused on discovering individual CA patterns, our implementation shifts to studying multiple interacting patterns potentially governed by different update rules. Rather than optimizing for interesting static properties, we use simulation-wide metrics related to evolutionary activity, compressibility and entropy at different levels of coarse-graining, which allows us to efficiently explore conditions that promote interesting long-term dynamics, rather than merely snapshots. 

We applied the approach to Flow-Lenia in two complementary settings: an open environment targeting ecosystem-level dynamics, and an environment filled with obstacles targeting matter movement. In both, IMGEP illuminated substantially more of the goal-space than random search and revealed qualitatively distinct self-organized behaviors, resembling various biological phenomena. Leveraging the resulting archive, a scaling study across six spatial scales and seven horizons then uncovered macro-scale coherent structures larger than entire base-sized grids.
The introduced interactive exploration tool proved highly valuable during analysis, allowing us to quickly traverse large datasets and identify interesting trials. It also helped us assess the usefulness of the proposed metrics. 

Our multi-metric evaluation framework suggested that the most interesting configurations were not those maximizing any single metric, but rather those exhibiting balance across multiple dimensions. Configurations showing sustained evolutionary activity often featured organization across different spatial scales, maintained parameter diversity despite matter exchange, and demonstrated dynamic yet persistent interactions. While qualitative, this observation aligns with theories that complex phenomena often emerge at the boundary between ordered and chaotic regimes. 

Beyond the specific discoveries, our experiments demonstrate a recipe for using the framework: diversity search on small-scale simulations can act as a principled scaffold for designing subsequent, large-scale experiments. The IMGEP archive provided a curated set of diverse universes from which farthest-point sampling selected configurations to rerun across spatial and temporal scales. Inspection of the scaled runs then surfaced new insights and questions: whether the macro-patterns persist or are transient, what are the properties of the macro-patterns, which metrics retain discriminative power at scale, and whether further scaling would yield yet richer phenomena. All of these in turn motivate the next round of search and metric design. We see this design–inspection–redesign cycle as the framework's most generalizable takeaway.

\subsection{Challenges and future work}
Future work falls into two complementary areas: improving the diversity search itself, and better understanding the dynamics Flow-Lenia produces.

\textbf{Improving the search.} Several refinements are worth pursuing. Our algorithm samples goals uniformly at random, but it would likely benefit from an automatically generated curriculum — for example, one that concentrates sampling along the frontier of the explored region. We also did not systematically investigate the known failure modes of diversity search: in high-dimensional goal-spaces, nearest-neighbor selection can repeatedly return to the same small set of trials, causing the search to stall. The per-dimension Gaussian noise mutation operator is similarly simple; tuning its noise scales or adopting more informed operators is a clear way to improve sample efficiency. We did not pursue these directions because goal-space coverage continued to grow even in long runs, we prioritized analyzing the discoveries we already had. Revisiting these design choices is a clear next step.

\textbf{Understanding Flow-Lenia.} Beyond goal-space and local-parameter statistics, our analysis of the resulting dynamics is largely qualitative. The macro-scale patterns uncovered in the scaling study deserve dedicated attention; how best to characterize their structure, lifetimes, and interactions remains an open question. We also observed that Flow-Lenia dynamics are highly sensitive: a single local-rule mutation or a small perturbation of the initial conditions can substantially alter a simulation's trajectory. Flow-Lenia exposes many parameters whose individual and joint effects are not well understood, and a fuller treatment lies beyond the scope of this paper. The mixing rules in particular merit closer study, as our 
preliminary study already suggests they play a central role.

Given our compute and time budget, we pushed the scale of the experiments to their limits. As with artificial intelligence, we believe some of artificial life's most exciting discoveries will come from scaling, and look forward to seeing what phenomena emerge within even larger Flow-Lenia universes.

Our investigation of the gathered experimental data has only scratched the surface. In addition to releasing our interactive tool, we will open-source the full codebase and generated dataset to facilitate reproducibility and encourage further research. We invite the community to use the released resources to design new metrics, craft richer environments, and develop new analytical methods for studying ecosystem dynamics in CAs and beyond. Applying the presented method and metrics to other substrates could help identify common conditions that foster ecological phenomena and, hopefully, contribute to the long quest of the Artificial Life community to bootstrap open-ended evolution in silico.

\printbibliography

@article{etcheverry2020hierarchically,
  title={Hierarchically organized latent modules for exploratory search in morphogenetic systems},
  author={Etcheverry, Mayalen and Moulin-Frier, Cl{\'e}ment and Oudeyer, Pierre-Yves},
  journal={Advances in Neural Information Processing Systems},
  volume={33},
  pages={4846--4859},
  year={2020}
}

@book{Mitchell2010Complexity,
  author    = {Mitchell, Melanie},
  title     = {Complexity: A Guided Tour},
  publisher = {Oxford University Press},
  address   = {New York, NY},
  year      = {2009},
  isbn      = {978-0-19-512441-5}
}

@phdthesis{etcheverry2023curiosity,
  title={Curiosity-driven AI for Science: Automated Discovery of Self-Organized Structures},
  author={Etcheverry, Mayalen},
  year={2023},
  school={Universit{\'e} de Bordeaux}
}

@inproceedings{faldor2024toward,
  title={Toward Artificial Open-Ended Evolution within Lenia using Quality-Diversity},
  author={Faldor, Maxence and Cully, Antoine},
  booktitle={ALIFE 2024: Proceedings of the 2024 Artificial Life Conference},
  year={2024},
  organization={MIT Press}
}

@article{cully2015robots,
  title={Robots that can adapt like animals},
  author={Cully, Antoine and Clune, Jeff and Tarapore, Danesh and Mouret, Jean-Baptiste},
  journal={Nature},
  volume={521},
  number={7553},
  pages={503--507},
  year={2015},
  publisher={Nature Publishing Group UK London}
}

@book{adamatzky2010game,
  title     = {Game of life cellular automata},
  author    = {Adamatzky, Andrew},
  volume    = {1},
  year      = {2010},
  publisher = {Springer}
}

@article{chan2018lenia,
  title   = {Lenia-biology of artificial life},
  author  = {Chan, Bert Wang-Chak},
  journal = {arXiv preprint arXiv:1812.05433},
  year    = {2018}
}

@article{hamon2024discovering,
  title   = {Discovering Sensorimotor Agency in Cellular Automata using Diversity Search},
  author  = {Hamon, Gautier and Etcheverry, Mayalen and Chan, Bert Wang-Chak and Moulin-Frier, Cl{\'e}ment and Oudeyer, Pierre-Yves},
  journal = {arXiv preprint arXiv:2402.10236},
  year    = {2024}
}

@inproceedings{hickinbotham2015conservation,
  title        = {Conservation of matter increases evolutionary activity},
  author       = {Hickinbotham, Simon John and Stepney, Susan},
  booktitle    = {European Conference on Artificial Life 2015},
  pages        = {98--105},
  year         = {2015},
  organization = {MIT Press}
}

@article{langton1984self,
  title     = {Self-reproduction in cellular automata},
  author    = {Langton, Christopher G},
  journal   = {Physica D: Nonlinear Phenomena},
  volume    = {10},
  number    = {1-2},
  pages     = {135--144},
  year      = {1984},
  publisher = {Elsevier}
}

@inproceedings{plantec2023flow,
  title        = {Flow-Lenia: Towards open-ended evolution in cellular automata through mass conservation and parameter localization},
  author       = {Plantec, Erwan and Hamon, Gautier and Etcheverry, Mayalen and Oudeyer, Pierre-Yves and Moulin-Frier, Cl{\'e}ment and Chan, Bert Wang-Chak},
  booktitle    = {Artificial Life Conference Proceedings 35},
  volume       = {2023},
  pages        = {131},
  year         = {2023},
  organization = {MIT Press}
}

@article{chan2023large,
  title     = {Towards {{Large-Scale Simulations}} of {{Open-Ended Evolution}} in {{Continuous Cellular Automata}}},
  author    = {Chan, Bert Wang-Chak},
  year      = {2023},
  journal   = {arXiv preprint arXiv:2304.05639}
}

@inproceedings{soros2014identifying,
  title        = {Identifying necessary conditions for open-ended evolution through the artificial life world of chromaria},
  author       = {Soros, Lisa and Stanley, Kenneth},
  booktitle    = {Artificial Life Conference Proceedings},
  pages        = {793--800},
  year         = {2014},
  organization = {MIT Press}
}

@article{stanley2019open,
  title     = {Why open-endedness matters},
  author    = {Stanley, Kenneth O},
  journal   = {Artificial life},
  volume    = {25},
  number    = {3},
  pages     = {232--235},
  year      = {2019},
  publisher = {MIT Press}
}

@article{taylor2015requirements,
  title   = {Requirements for open-ended evolution in natural and artificial systems},
  author  = {Taylor, Tim},
  journal = {arXiv preprint arXiv:1507.07403},
  year    = {2015}
}

@article{dolson2019modes,
  title     = {The MODES toolbox: Measurements of open-ended dynamics in evolving systems},
  author    = {Dolson, Emily L and Vostinar, Anya E and Wiser, Michael J and Ofria, Charles},
  journal   = {Artificial life},
  volume    = {25},
  number    = {1},
  pages     = {50--73},
  year      = {2019},
  publisher = {MIT Press}
}

@misc{reinke2020a,
  title     = {Intrinsically {{Motivated Discovery}} of {{Diverse Patterns}} in {{Self-Organizing Systems}}},
  author    = {Reinke, Chris and Etcheverry, Mayalen and Oudeyer, Pierre-Yves},
  year      = {2020},
  month     = feb,
  number    = {arXiv:1908.06663},
  publisher = {{arXiv}},
  doi       = {10.48550/arXiv.1908.06663}
}

@article{salzberg2004,
  title   = {Complex Genetic Evolution of Artificial Self-Replicators in Cellular Automata},
  author  = {Salzberg, Chris and Sayama, Hiroki},
  year    = {2004},
  journal = {Complexity},
  volume  = {10},
  number  = {2},
  pages   = {33--39},
  issn    = {1099-0526},
  doi     = {10.1002/cplx.20060}
}

@article{sayama1999,
  title   = {Toward the Realization of an Evolving Ecosystem on Cellular Automata},
  author  = {Sayama, Hiroki},
  year    = {1999},
  journal = {Proc. Fourth Int. Symp. Artificial Life and Robotics},
  pages   = {254--257}
}

@misc{bedau1996,
  title  = {Measurement of {{Evolutionary Activity}}, {{Teleology}}, and {{Life}}},
  author = {Bedau, Mark A. and Packard, Norman H.},
  year   = {1996}
}

@article{beer2004,
  title   = {Autopoiesis and {{Cognition}} in the {{Game}} of {{Life}}},
  author  = {Beer, Randall D.},
  year    = {2004},
  month   = jul,
  journal = {Artificial Life},
  volume  = {10},
  number  = {3},
  pages   = {309--326},
  issn    = {1064-5462},
  doi     = {10.1162/1064546041255539}
}

@inproceedings{droop2012,
  title     = {A Quantitative Measure of Non-Neutral Evolutionary Activity for Systems That Exhibit Intrinsic Fitness},
  booktitle = {{{ALIFE}} 2012: {{The Thirteenth International Conference}} on the {{Synthesis}} and {{Simulation}} of {{Living Systems}}},
  author    = {Droop, Alastair and Hickinbotham, Simon},
  year      = {2012},
  month     = jul,
  pages     = {45--52},
  publisher = {{MIT Press}},
  doi       = {10.1162/978-0-262-31050-5-ch007}
}

@incollection{lehman2011a,
  title     = {Novelty {{Search}} and the {{Problem}} with {{Objectives}}},
  booktitle = {Genetic {{Programming Theory}} and {{Practice IX}}},
  author    = {Lehman, Joel and Stanley, Kenneth O.},
  editor    = {Riolo, Rick and Vladislavleva, Ekaterina and Moore, Jason H.},
  year      = {2011},
  series    = {Genetic and {{Evolutionary Computation}}},
  pages     = {37--56},
  publisher = {{Springer}},
  address   = {{New York, NY}},
  doi       = {10.1007/978-1-4614-1770-5_3},
  isbn      = {978-1-4614-1770-5}
}

@inproceedings{cisneros2019evolving,
  title        = {Evolving structures in complex systems},
  author       = {Cisneros, Hugo and Sivic, Josef and Mikolov, Tomas},
  booktitle    = {2019 IEEE Symposium Series on Computational Intelligence (SSCI)},
  pages        = {230--237},
  year         = {2019},
  organization = {IEEE}
}

@book{langton1995artificial,
    author = {Langton, Christopher G.},
    title = {Artificial Life: An Overview},
    publisher = {The MIT Press},
    year = {1995},
    month = {07},
    isbn = {9780262277921},
    doi = {10.7551/mitpress/1427.001.0001},
}

@article{Beer2014Cognitive,
    author = {Beer, Randall D.},
    title = {The Cognitive Domain of a Glider in the Game of Life},
    journal = {Artificial Life},
    volume = {20},
    number = {2},
    pages = {183-206},
    year = {2014},
    month = {04},
    issn = {1064-5462},
    doi = {10.1162/ARTL_a_00125},
    eprint = {https://direct.mit.edu/artl/article-pdf/20/2/183/1664518/artl\_a\_00125.pdf},
}

@article{zenil2015two,
  title     = {Two-dimensional Kolmogorov complexity and an empirical validation of the Coding theorem method by compressibility},
  author    = {Zenil, Hector and Soler-Toscano, Fernando and Delahaye, Jean-Paul and Gauvrit, Nicolas},
  journal   = {PeerJ Computer Science},
  volume    = {1},
  pages     = {e23},
  year      = {2015},
  publisher = {PeerJ Inc.}
}

@article{lloyd1988complexity,
  title     = {Complexity as thermodynamic depth},
  author    = {Lloyd, Seth and Pagels, Heinz},
  journal   = {Annals of physics},
  volume    = {188},
  number    = {1},
  pages     = {186--213},
  year      = {1988},
  publisher = {Elsevier}
}

@article{langton1990computation,
  title     = {Computation at the edge of chaos: Phase transitions and emergent computation},
  author    = {Langton, Chris G},
  journal   = {Physica D: nonlinear phenomena},
  volume    = {42},
  number    = {1-3},
  pages     = {12--37},
  year      = {1990},
  publisher = {Elsevier}
}

@article{colas2022autotelic,
  title={Autotelic agents with intrinsically motivated goal-conditioned reinforcement learning: a short survey},
  author={Colas, C{\'e}dric and Karch, Tristan and Sigaud, Olivier and Oudeyer, Pierre-Yves},
  journal={Journal of Artificial Intelligence Research},
  volume={74},
  pages={1159--1199},
  year={2022}
}

@article{wolfram1984universality,
  title     = {Universality and complexity in cellular automata},
  author    = {Wolfram, Stephen},
  journal   = {Physica D: Nonlinear Phenomena},
  volume    = {10},
  number    = {1-2},
  pages     = {1--35},
  year      = {1984},
  publisher = {Elsevier}
}

@article{oudeyer2007intrinsic,
  title     = {Intrinsic motivation systems for autonomous mental development},
  author    = {Oudeyer, Pierre-Yves and Kaplan, Frdric and Hafner, Verena V},
  journal   = {IEEE transactions on evolutionary computation},
  volume    = {11},
  number    = {2},
  pages     = {265--286},
  year      = {2007},
  publisher = {IEEE}
}

@article{etcheverry2024ai,
  title     = {AI-driven Automated Discovery Tools Reveal Diverse Behavioral Competencies of Biological Networks},
  author    = {Etcheverry, Mayalen and Moulin-Frier, Cl{\'e}ment and Oudeyer, Pierre-Yves and Levin, Michael},
  journal   = {eLife},
  volume    = {13},
  year      = {2024},
  publisher = {eLife Sciences Publications Limited}
}

@article{forestier2022intrinsically,
  title   = {Intrinsically motivated goal exploration processes with automatic curriculum learning},
  author  = {Forestier, S{\'e}bastien and Portelas, R{\'e}my and Mollard, Yoan and Oudeyer, Pierre-Yves},
  journal = {Journal of Machine Learning Research},
  volume  = {23},
  number  = {152},
  pages   = {1--41},
  year    = {2022}
}

@article{gottweis2025towards,
  title={Towards an AI co-scientist},
  author={Gottweis, Juraj and Weng, Wei-Hung and Daryin, Alexander and Tu, Tao and Palepu, Anil and Sirkovic, Petar and Myaskovsky, Artiom and Weissenberger, Felix and Rong, Keran and Tanno, Ryutaro and others},
  journal={arXiv preprint arXiv:2502.18864},
  year={2025}
}

@article{kumar2024automating,
  title={Automating the Search for Artificial Life with Foundation Models},
  author={Kumar, Akarsh and Lu, Chris and Kirsch, Louis and Tang, Yujin and Stanley, Kenneth O and Isola, Phillip and Ha, David},
  journal={arXiv preprint arXiv:2412.17799},
  year={2024}
}

@article{grizou2020curious,
  title={A curious formulation robot enables the discovery of a novel protocell behavior},
  author={Grizou, Jonathan and Points, Laurie J and Sharma, Abhishek and Cronin, Leroy},
  journal={Science advances},
  volume={6},
  number={5},
  pages={eaay4237},
  year={2020},
  publisher={American Association for the Advancement of Science}
}

@article{falk2024curiosity,
  title={Curiosity-driven search for novel nonequilibrium behaviors},
  author={Falk, Martin J and Roach, Finnegan D and Gilpin, William and Murugan, Arvind},
  journal={Physical Review Research},
  volume={6},
  number={3},
  pages={033052},
  year={2024},
  publisher={APS}
}

@article{lu2024ai,
  title={The ai scientist: Towards fully automated open-ended scientific discovery},
  author={Lu, Chris and Lu, Cong and Lange, Robert Tjarko and Foerster, Jakob and Clune, Jeff and Ha, David},
  journal={arXiv preprint arXiv:2408.06292},
  year={2024}
}

@article{faldor2024omni,
  title={OMNI-EPIC: Open-endedness via Models of human Notions of Interestingness with Environments Programmed in Code},
  author={Faldor, Maxence and Zhang, Jenny and Cully, Antoine and Clune, Jeff},
  journal={arXiv preprint arXiv:2405.15568},
  year={2024}
}

@article{wang2023voyager,
  title={Voyager: An open-ended embodied agent with large language models},
  author={Wang, Guanzhi and Xie, Yuqi and Jiang, Yunfan and Mandlekar, Ajay and Xiao, Chaowei and Zhu, Yuke and Fan, Linxi and Anandkumar, Anima},
  journal={arXiv preprint arXiv:2305.16291},
  year={2023}
}

@article{pourcel2024aces,
  title={ACES: Generating a Diversity of Challenging Programming Puzzles with Autotelic Generative Models},
  author={Pourcel, Julien and Colas, C{\'e}dric and Molinaro, Gaia and Oudeyer, Pierre-Yves and Teodorescu, Laetitia},
  journal={Advances in Neural Information Processing Systems},
  volume={37},
  pages={67627--67662},
  year={2024}
}

@misc{moroz2020reintegration,
  author       = {Moroz, Mykhailo},
  title        = {Reintegration Tracking},
  year         = {2020},
  howpublished = {\url{https://michaelmoroz.github.io/Reintegration-Tracking/}},
  note         = {Accessed: 2025-05-05}
}

\end{document}